\documentclass[10pt,journal,compsoc]{IEEEtran}
\usepackage{amsmath,amsfonts}
\usepackage{algorithmic}
\usepackage{algorithm}
\usepackage{array}
\usepackage{textcomp}
\usepackage{subcaption}
\usepackage{stfloats}
\usepackage{url}
\usepackage{verbatim}
\usepackage{graphicx}
\usepackage{cite}
\usepackage[dvipsnames]{xcolor}
\usepackage{etoolbox}
\usepackage{longtable}
\usepackage{pdflscape}
\usepackage{graphicx}
\usepackage{siunitx}
\usepackage{subcaption}
\usepackage{booktabs}
\usepackage{amssymb}
\usepackage{utfsym}
\usepackage{colortbl}%
\usepackage[pagebackref=true,breaklinks=true,colorlinks,bookmarks=false,citecolor=blue,linkcolor=blue,urlcolor=blue]{hyperref}
  \newcommand{\myrowcolour}{\rowcolor[gray]{0.925}}
  
  \newcommand{\highest}[1]{\textcolor{Maroon}{\textbf{#1}}}

\usepackage[dvipsnames]{xcolor}
\usepackage{subcaption}
\usepackage[subfigure]{tocloft}

\newcommand{\ourmethod}{DrivingGaussian++}

\newif\ifdrafting
\draftingtrue 
\ifdrafting
    \newcommand{\todo}[1]{{\leavevmode\color[rgb]{1,0,0}[TODO: #1]}}
    
    \newcommand{\ds}[1]{{\leavevmode\color[rgb]{0,0,1}[DS: #1]}}

    \definecolor{tabfirst}{rgb}{1, 0.7, 0.7} 
    \definecolor{tabsecond}{rgb}{1, 0.85, 0.7} 
    \definecolor{tabthird}{rgb}{1, 1, 0.7} 
\else
    \newcommand{\todo}[1]{}
    \newcommand{\ds}[1]{}
    \newcommand{\mh}[1]{}
    
\fi

\newcommand{\aftertab}{\vspace{-1em}}
\newcommand{\afterfig}{\vspace{-1.25em}}
\newcommand{\aroundeqn}{\vspace{-.2em}}

\begin{document}

\title{\ourmethod{}: Towards Realistic Reconstruction and Editable Simulation for Surrounding Dynamic Driving Scenes}

\author{
Yajiao Xiong\textsuperscript{\S}, 
Xiaoyu Zhou\textsuperscript{\S},
Yongtao Wang\textsuperscript{*}, 
Deqing Sun\textsuperscript{\ddag}, 
Ming-Hsuan Yang\textsuperscript{\ddag}

\IEEEcompsocitemizethanks{
\IEEEcompsocthanksitem Preliminary work DrivingGaussian has been published in CVPR 2024. (Corresponding author: Yongtao Wang.)
\IEEEcompsocthanksitem Xiaoyu Zhou, Yajiao Xiong, and Yongtao Wang are with Wangxuan Institute of Computer Technology, Peking University (e-mail: xyrain.zhou@gmail.com, asdlkj@stu.pku.edu.cn, wyt@pku.edu.cn).

Deqing Sun is with Google DeepMind (deqing.sun@gmail.com).

Ming-Hsuan Yang is with Google DeepMind, and also with University of California, Merced (myang37@ucmerced.edu).

\IEEEcompsocthanksitem
$\S$: Equal contributions; $*$: Corresponding author; $\ddag$: Co-last authors.
}
}

\IEEEtitleabstractindextext{%
\begin{abstract}
We present \ourmethod{}, an efficient and effective framework for realistic reconstructing and controllable editing of surrounding dynamic autonomous driving scenes. 
\ourmethod{} models the static background using incremental 3D Gaussians and reconstructs moving objects with a composite dynamic Gaussian graph, ensuring accurate positions and occlusions. By integrating a LiDAR prior, it achieves detailed and consistent scene reconstruction, outperforming existing methods in dynamic scene reconstruction and photorealistic surround-view synthesis. \ourmethod{} supports training-free controllable editing for dynamic driving scenes, including texture modification, weather simulation, and object manipulation, leveraging multi-view images and depth priors. By integrating large language models (LLMs) and controllable editing, our method can automatically generate dynamic object motion trajectories and enhance their realism during the optimization process. \ourmethod{} demonstrates consistent and realistic editing results and generates dynamic multi-view driving scenarios, while significantly enhancing scene diversity. More results and code can be found at the project site: \href{https://xiong-creator.github.io/DrivingGaussian_plus.github.io/}{\text{https://drivinggaussian-plus.github.io/DrivingGaussian\_plus.github.io}}

\end{abstract}

}

\maketitle
\IEEEdisplaynontitleabstractindextext

\IEEEpeerreviewmaketitle

\IEEEraisesectionheading{\section{Introduction}}
\IEEEPARstart{W}{ith} the rapid advances of autonomous driving technologies, the availability of driving data has introduced new opportunities to enhance autonomous driving simulation. These datasets encapsulate rich semantic and spatial information, significantly improving the performance of downstream perception tasks. Among these, 3D scene editing is crucial in enhancing the data of autonomous driving. It facilitates the generation of diverse driving scenarios, which is particularly valuable in synthesizing complex and rare cases critical for testing and validation. Using 3D scene editing, it becomes possible to simulate diverse real-world driving conditions, thereby enhancing the robustness and safety of autonomous driving systems.

 \begin{figure}[t]
  \centering
  \vspace{-5mm}
  \includegraphics[width=0.9\linewidth]{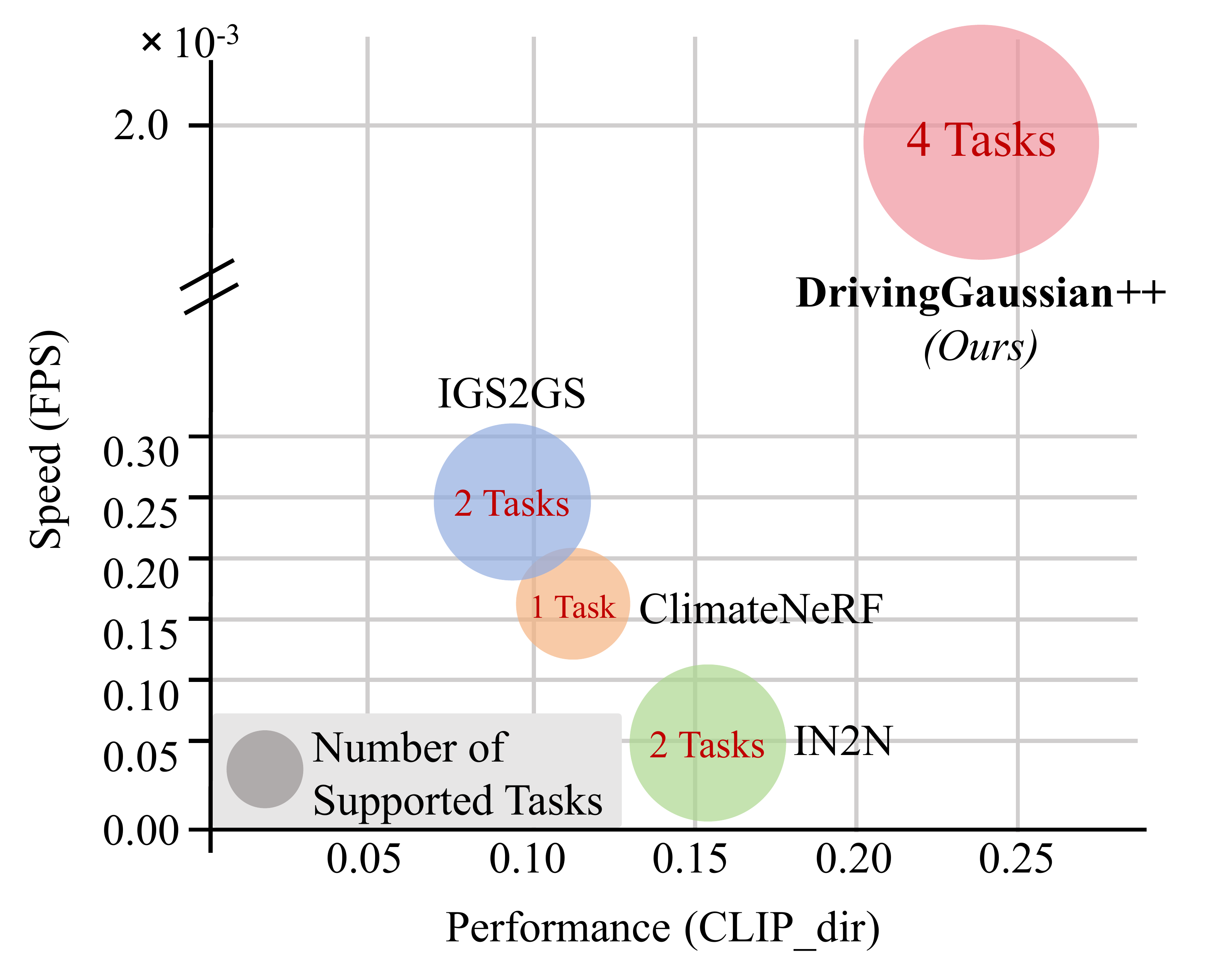}
  \caption{\textbf{\ourmethod{} achieves favorable trade-offs between speed, accuracy, and task diversity.} Our method enables fast, training-free execution of various 3D controllable editing tasks while delivering comparable or better performance to state-of-the-art task-specific models.}
  \label{fig:fig0}
  \afterfig
 \end{figure}

The 3D scene editing task encompasses a variety of components, including style transfer, motion modification, weather simulation, and the addition or removal of objects. However, due to the inherent diversity and complexity of these tasks, existing 3D scene editing methods~\cite{haque2023instruct,khandelwal2023infusion,wu2024gaussctrl,chen2024gaussianeditor,dong2024vica,li2023climatenerf} are often highly specialized, lacking a framework capable of comprehensively addressing multiple editing tasks. While the widely adopted strategy of leveraging 2D editing techniques achieves superior results, it requires repeated 2D editing during training to maintain multi-view consistency. This significantly increases computational costs, making it impractical for large-scale autonomous driving scenarios.

Representing and modeling large-scale dynamic scenes is the foundation for 3D scene editing and contributes to a series of autonomous driving tasks. Unfortunately, reconstructing complex 3D scenes from sparse vehicle-mounted sensor data is challenging, particularly when the ego vehicle is moving at high speeds. Furthermore, it becomes even more challenging in multi-camera settings due to their outward views, minimal overlaps, and variations in light from different directions. Complex geometry, diverse optical degradation, and spatiotemporal inconsistency also pose significant challenges in modeling such a 360-degree, large-scale driving scene.

Neural radiance fields~\cite{mildenhall2021nerf} (NeRFs) have recently emerged as a promising method for modeling scenery at the object or room level. 
Some recent studies~\cite{tancik2022block,turki2022mega,wang2023neural,zhenxing2022switch} have extended NeRF to large-scale, unbounded static scenes and multiple dynamic objects~\cite{ost2021neural,song2022towards}. 
However, NeRF-based methods are computationally intensive and require densely overlapping views and consistent lighting. These issues affect their ability to construct driving scenes with outward multi-camera setups at high speeds. Furthermore, network capacity limitations make it challenging for them to model long-term, dynamic scenes with multiple objects, leading to visual artifacts and blurring. 
The 3D Gaussian Splatting (3DGS)~\cite{kerbl20233d} methods represent scenes with explicit 3D Gaussian representation and achieve state-of-the-art performance in novel view synthesis. However, the original 3DGS still encounters significant challenges in modeling large-scale dynamic driving scenes due to fixed Gaussians and constrained representation capacity. 

In this paper, our method is based on a framework that represents the surrounding and dynamic driving scenes. 
The key idea is to hierarchically model the complex driving scene using sequential data from multiple sensors. We adopt Composite Gaussian Splatting to decompose the scene into static background and dynamic objects, reconstructing each part separately. Based on these, global rendering via Gaussian Splatting captures occlusion in the real world, encompassing static backgrounds and dynamic objects. Furthermore, we incorporate a LiDAR prior into the Gaussian representation, which enables the recovery of more precise geometry and maintains better multi-view consistency. 

\begin{figure}[t]
  \centering
  \includegraphics[width=0.9\linewidth]{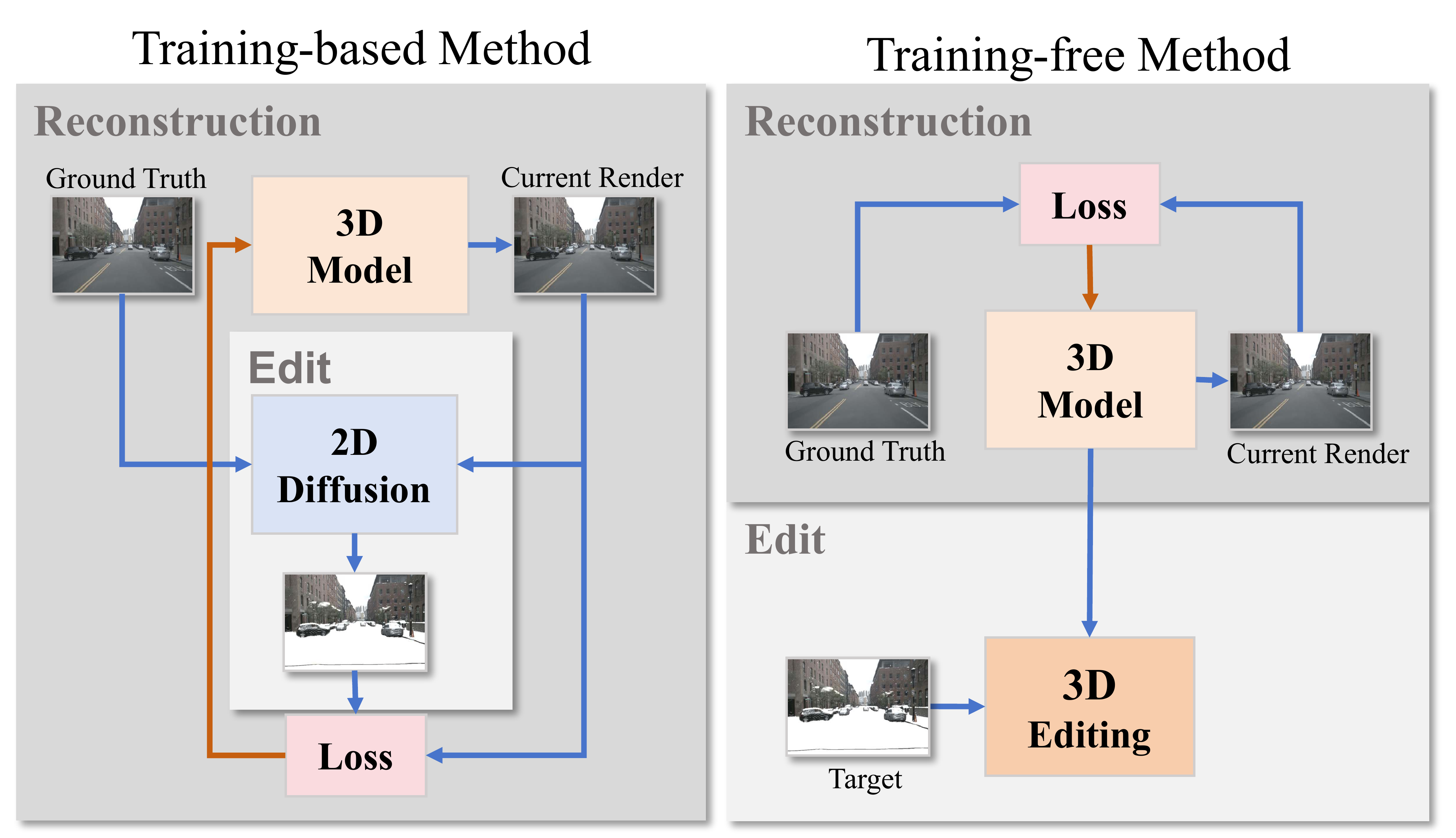}
  \caption{\textbf{Comparison between the traditional and our editing strategy.} Our method differs from previous work in three key aspects: 1) supporting diverse multi-task 3D simulation; 2) enabling training-free controllable editing; 3) demonstrating strong performance in complex dynamic driving environments.
   }
  \label{fig:fig1}
  \afterfig
 \end{figure}

We also introduce a multitask editing framework for autonomous driving scenes. Within this framework, we meticulously implement three key tasks: texture modification, weather simulation, and object manipulation. To mitigate computational and temporal costs, we propose the post-editing strategy, which effectively decouples the reconstruction and editing processes, as illustrated in Figure~\ref{fig:fig1}. We compare our method with IN2N~\cite{haque2023instruct}, IGS2GS~\cite{vachhainstruct}, and ClimateNeRF~\cite{li2023climatenerf}, which demonstrate superior processing speed and competitive performance. 
This approach enables editing without additional training, significantly reducing the overall training overhead. \ourmethod{} can be seamlessly applied to any pre-reconstructed 3D scene. Specifically, we employ Composite Gaussian Splatting to construct explicit scene representations using Gaussian distributions. By identifying and editing these Gaussians at the 3D level, we circumvent the issue of multi-view inconsistency that often arises when 2D models are directly applied to 3D spaces. Furthermore, we enhance the editing results by integrating advanced image-processing tools, demonstrating the potential of cross-dimensional technological integration in achieving high-quality scene modifications.

\vspace{1mm}
\noindent \textbf{Differences with preliminary results published in CVPR 2024}. We have extended our work in several aspects: (i) We represent large-scale, dynamic driving scenes based on Composite Gaussian Splatting, which introduces two novel modules, including Incremental Static 3D Gaussians and Composite Dynamic Gaussian Graphs. The former reconstructs the static background incrementally, while the latter models multiple dynamic objects with a Gaussian graph. (ii) We construct a scene editing framework to edit the reconstructed scene in a training-free manner, covering multiple tasks including texture modification, weather simulation, and object manipulation. It contributes to generating novel and realistic simulation data. (iii) We facilitate dynamic editing of driving scenes, which predicts the motion trajectory of particles inserted into the scene. (iv) 
We construct a foreground asset bank with 3D generation and reconstruction and validate the quality of the data.

\section{Related Work}
\subsection{3D Reconstruction}
\subsubsection{Neural Radiance Fields}
Rapid progress in neural rendering for novel view synthesis has received significant attention. Neural Radiance Fields (NeRFs), which utilizes multi-layer perceptrons (MLPs) and differentiable volume rendering, can reconstruct 3D scenes and synthesize novel views from a set of 2D images and corresponding camera pose information.

\noindent \textbf{NeRF for Bounded Scene.} 
Typical NeRF models are limited to bounded scenes, requiring a consistent distance between the center object and the camera. NeRF models also struggle with scenes captured with slight overlaps and outward capture methods.
Numerous advances have expanded the capabilities of NeRFs, leading to improvements in training speed~\cite{muller2022instant, fridovich2022plenoxels, garbin2021fastnerf}, pose optimization~\cite{lin2021barf, wang2021nerf, bian2023nope}, scene editing~\cite{rudnev2022nerf, li2023climatenerf}, and dynamic scene representation~\cite{pumarola2021d, huang2022hdr}.
However, applying NeRF to large-scale unbounded scenes, such as autonomous driving scenarios, remains challenging.

\noindent \textbf{NeRF for Unbounded Scenes.} For large-scale unbounded scenes, numerous methods~\cite{tancik2022block, turki2022mega, zhenxing2022switch, wang2023neural, martin2021nerf} have introduced refined versions of NeRF to model multi-scale urban-level static scenes. 
Inspired by the mipmapping approach to preventing aliasing, MIP-NeRF models~\cite{barron2021mip, barron2022mip} have been developed to account for unbounded scenes. 
To enable high-fidelity rendering, Xu et al.~\cite{xu2023grid} combine the compact multi-resolution ground feature planes with NeRF for large urban scenes.
On the other hand, Guo et al.~\cite{guo2023streetsurf} propose a disentanglement approach that can model unbounded street views but ignores dynamic objects on the road.
However, these methods model scenes under the assumption that the scene remains static and face challenges in effectively capturing dynamic elements.

Although most existing NeRF-based methods rely on accurate camera poses, several approaches have been developed~\cite{liu2023robust, meuleman2023progressively} to synthesize the view of dynamic monocular videos. 
However, these methods are limited to forward monocular views and are unable to effectively handle inputs from surrounding multi-camera setups. 
For dynamic urban scenes, Ost et al.~\cite{ost2021neural, song2022towards} extend NeRF to dynamic scenes with multiple objects using a scene graph, and Wu et al.~\cite{wu2023mars, yang2023unisim} propose instance-aware realistic simulators for monocular dynamic scenes. 
On the other hand, Xie et al.~\cite{xie2023s} improve the parameterization and camera poses of the surrounding views while using LiDAR as additional depth supervision. In addition, several methods~\cite{turki2023suds, yang2023emernerf} decompose the scene into static background and dynamic objects and construct the scene with the help of LiDAR and 2D optical flow.

The quality of views synthesized by NeRF-based methods deteriorates in scenarios with multiple dynamic objects and variations and lighting variations, owing to their dependency on ray sampling. In addition, the utilization of LiDAR is confined to providing auxiliary depth supervision, and its potential benefits in reconstruction, such as delivering geometric priors, have not been explored.

To address these limitations, we utilize Composite Gaussian Splatting to model the unbounded dynamic scenes, where the static background is incrementally reconstructed as the ego vehicle moves, and multiple dynamic objects are modeled and integrated into the entire scene through Gaussian graphs.
LiDAR is employed as the initialization for Gaussians, providing a more accurate geometric shape prior and a comprehensive scene description rather than solely serving as depth supervision for images.

\subsubsection{3D Gaussian Splatting}
The recent 3D Gaussian Splatting method~\cite{kerbl20233d} represents a static scene with numerous 3D Gaussians and achieves state-of-the-art results in novel view synthesis and training speeds. Compared with existing explicit scene representations (e.g., mesh, voxels), the 3DGS can model complex shapes with fewer parameters. 
Unlike implicit neural rendering, the 3DGS allows fast rendering and differentiable computation with splat-based rasterization.

\noindent \textbf{Dynamic 3D Gaussian Splatting.} While the original 3DGS is designed to represent static scenes, several methods have been developed for dynamic objects/scenes. Given a set of dynamic monocular images, Yang et al.~\cite{yang2023deformable} introduce a deformation network to model the motion of Gaussians. In addition, Wu et al.~\cite{wu20234d} connect adjacent Gaussians via a HexPlane, enabling real-time rendering. However, these two approaches are explicitly designed for monocular single-camera scenes focused on a central object. 
Luiten et al.~\cite{luiten2023dynamic} parameterize the entire scene using a set of dynamic Gaussians that evolve. However, it requires a camera array with dense multi-view images as inputs.

In real-world autonomous driving scenes, the high-speed movement of data collection platforms leads to extensive and complex background variations, often captured by sparse views (e.g., 2-4 views). Moreover, fast-moving dynamic objects with intense spatial changes and occlusion further complicate the situation. Collectively, these factors pose significant challenges for existing methods.

\subsection{3D Scene Cotrollable Editing}

Representing and modeling 3D scenes is fundamental for novel view synthesis. Neural Radiation Field (NeRF) and 3D Gaussian Splatting are two prominent methods for 3D scene reconstruction. NeRF implicitly encodes scene geometry and appearance within a multilayer perceptron (MLP), while 3D Gaussian Splatting explicitly represents the scene using 3D Gaussian ellipsoids. Despite the demonstrated reconstruction capabilities, editing these representations remains a significant challenge. Current approaches can be broadly classified into two categories: editing guided by diffusion models and editing based on 3D particle systems.

\subsubsection{Editing Based on Diffusion Guidance}
Diffusion models have gained much attention for their ability to enable text-driven image editing. Recent methods~\cite{haque2023instruct,dong2024vica,wu2024gaussctrl,zhuang2023dreameditor,zhu2023hifa,chen2024gaussianeditor} have extended this capability to 3D scenes using pre-trained diffusion models. These methods add noise to images rendered from 3D models. The noisy images, augmented with additional control conditions, are processed by a 2D diffusion model, which predicts the noise that represents the discrepancy between the target output and the input. This predicted noise is then used to compute the Score Distillation Sampling (SDS) loss, guiding the optimization of the 3D model. Despite their effectiveness, these approaches struggle with maintaining multi-view consistency and managing complex scene dynamics, especially in large-scale environments. Instruct-NeRF2NeRF (IN2N)~\cite{haque2023instruct} uses text commands to edit 3D models by transforming the 3D editing task into a 2D image editing problem. 
 
However, due to the inability to ensure consistent editing across surrounding views, this approach suffers from instability, slow processing speeds, and noticeable artifacts, particularly in 360-degree scenes. ViCA-NeRF~\cite{dong2024vica} adopts a similar strategy to IN2N, selecting a subset of reference images from the scene dataset and editing them, while blending the remaining images based on the projection results. Although this blending approach mitigates some issues, it does not resolve consistency problems and often results in blurry edits. 

Recently, DreamEditor~\cite{zhuang2023dreameditor} converts NeRF representations into mesh surfaces and directly optimizes the mesh using SDS loss and DreamBooth. HiFA~\cite{zhu2023hifa} improves multi-view consistency by dynamically adjusting the diffusion timestep and reducing the weight of the noise signal. Gaussian-based approaches~\cite{wu2024gaussctrl, chen2024gaussianeditor} extend these NeRF editing techniques to 3D Gaussian Splatting, leveraging depth models to estimate image depth as a geometric prior for the conditional diffusion model. 
Although these methods achieve more consistent 3D editing, they are still limited to texture modification due to the fixed nature of depth estimation. Additionally, most of them rely on static 2D and 3D masks to constrain editing regions, which are ineffective for dynamic 3D model training. These approaches have been primarily validated on object-centric datasets and remain unexplored in complex driving scenes.

In contrast, \ourmethod{} employs a training-free paradigm to effectively address the challenges of existing methods in dynamic driving scene editing, achieving exceptional editing consistency and visual quality.

\subsubsection{Editing Based on 3D Particle Systems.}
In addition to text-driven editing guided by diffusion models, numerous domain-specific editing methods operate without relying on additional target images~\cite{li2023climatenerf,khandelwal2023infusion}. ClimateNeRF~\cite{li2023climatenerf} simulates particle-based entities for various weather conditions, such as snow, fog, and floods, and integrates them into the original neural field to achieve realistic weather simulation. Similarly, GaussianEditor and Infusion~\cite{khandelwal2023infusion} leverage 3D Gaussian Splatting for controlled editing of 3D scenes. GaussianEditor identifies editing regions by training Gaussians with semantic attributes, enabling precise insertion and deletion at the 3D level. Infusion employs a depth completion model to establish depth information, which serves as a control signal for Gaussian completion. These methods demonstrate higher editing efficiency and superior multi-view consistency compared to diffusion model-guided approaches. 
Inspired by the recent advances, \ourmethod{} adopts 3D particle-level editing and further extends it to multiple tasks, including texture, object, and weather editing. 
Through a training-free paradigm, our approach achieves explicit, controllable, and efficient editing for large-scale autonomous driving scenes.

\begin{figure*}[ht]
  \centering
  \includegraphics[width=0.85\linewidth]{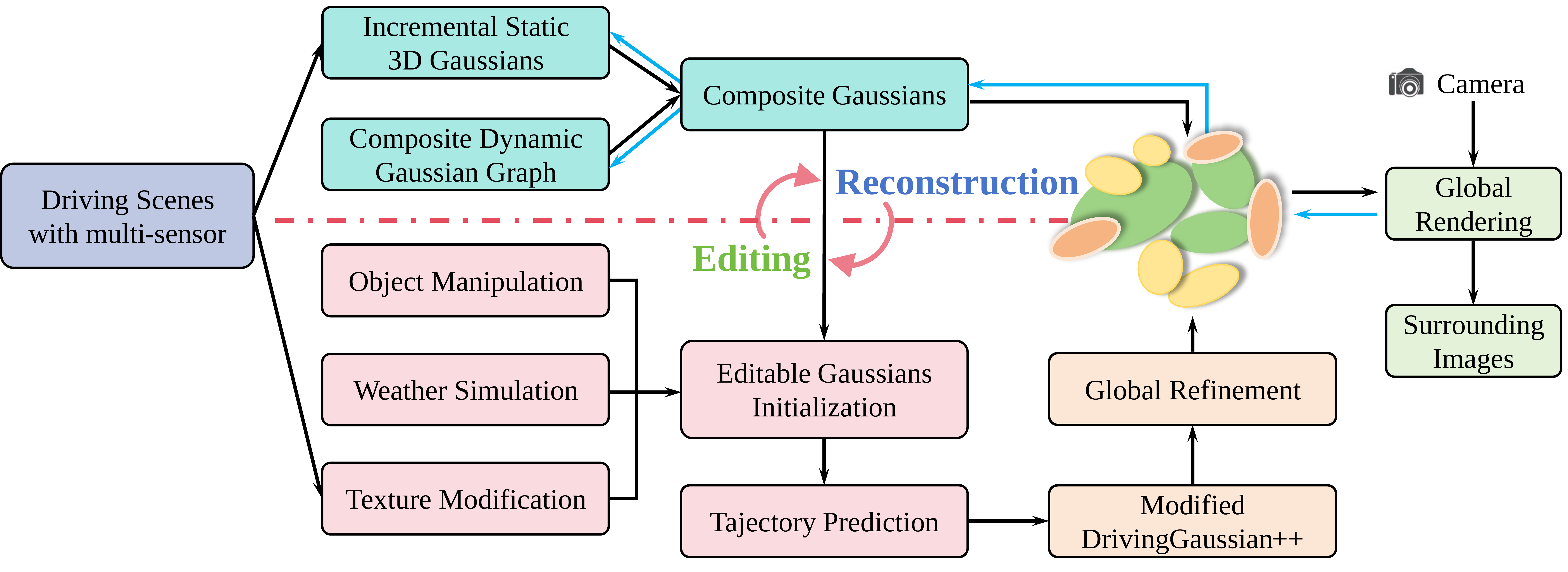}
  \caption{\textbf{Overview Pipeline of our method.} \ourmethod{} facilitates the reconstruction and controllable editing of surrounding dynamic scenes in autonomous driving by leveraging a compositional, controllable 3D representation with a unified global optimization strategy.}
  \label{fig:overview}
  \afterfig
\end{figure*}

\section{Method}

\begin{figure*}[tp]
  \centering
  \includegraphics[width=\linewidth]{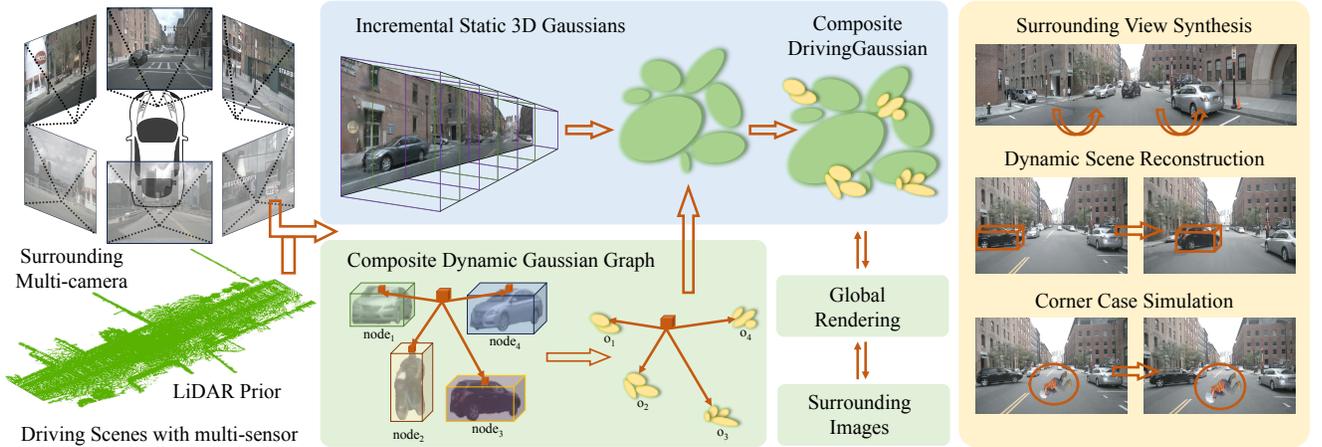}
  \caption{\textbf{Pipeline of our reconstruction method.} \textbf{Left:} \ourmethod{} takes sequential data from multi-sensor, including multi-camera images and LiDAR, as input. \textbf{Middle:} To represent large-scale dynamic driving scenes, we propose Composite Gaussian Splatting, which consists of two components. The first part incrementally reconstructs the extensive static background, while the second constructs multiple dynamic objects with a Gaussian graph and dynamically integrates them into the scene. \textbf{Right:} \ourmethod{} demonstrates good performance across multiple tasks and downstream applications.}
  \label{fig:rec_overview}
  \afterfig
\end{figure*}

Our goal is to achieve training-free editing within 3D autonomous driving scenes. To address multiple editing tasks, we propose a controllable and efficient framework. First, we accurately reconstruct dynamic driving scenes employing Composite Gaussian Splatting. Next, we identify specific Gaussians within the scene for modification or generate new Gaussians to simulate specific physical entities. These targeted Gaussians are then integrated into the original scene, where we predict the future trajectories of the objects. Finally, we refine the results using image processing techniques to enhance realism. Using this framework, we develop detailed editing methodologies for three key tasks: texture modification, weather simulation, and object manipulation. Our method is described in Fig.~\ref{fig:overview}

\subsection{Composite Gaussian Splatting}

3DGS performs well in static scenes, but has significant limitations in mixed scenes involving large-scale static backgrounds and multiple dynamic objects. 
As illustrated in Figure~\ref{fig:rec_overview}, our objective is to represent surrounding large-scale driving scenes with Composite Gaussian Splatting for unbounded static backgrounds and dynamic objects.

\subsubsection{LiDAR Prior with surrounding views}
\label{LiDAR}
The primitive 3DGS attempts to initialize Gaussians via structure-from-motion (SfM). However, unbounded urban scenes for autonomous driving contain many multi-scale backgrounds and foregrounds. Nevertheless, they are only glimpsed through exceedingly sparse views, resulting in erroneous and incomplete recovery of geometric structures.

To provide better initialization for Gaussians, we introduce the LiDAR prior to 3D Gaussian to obtain better geometries and maintain multi-camera consistency in surrounding view registration. At each timestep $t \in T$, given a set of multi-camera images $ \{ I_{t}^{i} | i = 1 \ldots N \}$ collected from the moving platform and multi-frame LiDAR sweeps ${L_{t}}$. 
We minimize multi-camera registration errors using LiDAR-image multi-modal data and obtain accurate point positions and geometric priors.

We first merge multiple frames of LiDAR sweeps to obtain the complete point cloud of the scene, denoted as $L$. We follow Colmap~\cite{schonberger2016structure} and extract image features $X = {x_p^q}$ from each image individually. Next, we project the LiDAR points onto the surrounding images. For each LiDAR point $l$, we transform its coordinates to the camera coordinate system and match it with the 2D-pixel of the camera image plane through projection:
\aroundeqn
\begin{equation}
  \mathbf{x_p^q} = \mathbf{K} [\mathbf{R_t^{i}} \cdot l_s + \mathbf{T_t^{i}}] \,,
  \label{eq:project}
\end{equation}
\aroundeqn
where $\mathbf{x_p^q}$ is the 2D pixel of the image, $\mathbf{I_{t}^{i}}$, $\mathbf{R_t^{i}}$ and $\mathbf{T_t^{i}}$ are orthogonal rotation matrices and translation vectors, respectively. In addition, $\mathbf{K} \in \mathbb{R}^{3 \times 3}$ represents the known camera intrinsic parameters. Notably, points from LiDAR might be projected onto multiple pixels across multiple images. Therefore, we select the point with the shortest Euclidean distance to the image plane and retain it as the projected point, assigning color.

Similar to existing 3D reconstruction methods~\cite{schmied2023r3d3, herau2023moisst}, we extend the dense bundle adjustment (DBA) to a multi-camera setup and obtain the updated LiDAR points. Experimental results show that initializing with LiDAR prior to aligning with surrounding multi-camera aids in providing the Gaussian model with more precise geometry priors.

\subsubsection{Incremental Static 3D Gaussians}
\label{static}

The static backgrounds of driving scenes pose challenges for scene modeling and editing due to their large scale, long duration, and variations caused by ego vehicle movement with multi-camera transformation.
As a vehicle moves, the static background frequently undergoes temporal shifts and dynamic changes. Due to the perspective principle, prematurely incorporating distant street scenes from time steps far away from the current can lead to scale confusion, resulting in unpleasant artifacts and blurring. 
To address this issue, we improve 3DGS by introducing Incremental Static 3D Gaussians, leveraging the perspective changes introduced by the vehicle's movement and the temporal relationships between adjacent frames, as shown in Figure~\ref{static-dynamic}.

We uniformly divide the static scene into $N$ bins based on the depth range provided by the LiDAR prior (Section~\ref{LiDAR}). These bins are arranged in chronological order, denoted as $\{\textrm{b}_i\}^{N}$, where each bin contains multi-camera images from one or more time steps. Neighboring bins have a small overlap region, which is used to align the static backgrounds of two bins. The latter bin is then incrementally fused into the Gaussian field of the previous bins.
For the scene within the first bin, we initialize the Gaussian model using the LiDAR prior (similarly applicable to SfM points):
\aroundeqn
\begin{equation}
  p_{b_0}(l | \mathbf{\mu}, \mathbf{\Sigma}) = e^{-\frac{1}{2}(\mathbf{l}-\mathbf{\mu})^{\top} \mathbf{\Sigma}^{-1}(\mathbf{l}-\mathbf{\mu})} \,,
\end{equation}
\aroundeqn
where $\mathbf{l} \in \mathbb{R}^{3}$ is the position of the LiDAR prior; $\mathbf{\mu}$ is the mean of the LiDAR points; $\mathbf{\Sigma} \in \mathbb{R}^{3 \times 3}$ is an anisotropic covariance matrix; and $^\top$ is the transpose operator. 
We utilize the surrounding views within this bin segment as supervision to update the parameters of the Gaussian model, including position $P(x,y,z)$, covariance matrix $\mathbf{\Sigma}$, coefficients of spherical harmonics for view-dependent color $C(r,g,b)$, along with an opacity $\alpha$.

For the subsequent bins, we use the Gaussians from the previous bin as the position priors and align the adjacent bins based on their overlapping regions. The 3D center for each bin can be defined as:
\aroundeqn
\begin{equation}
  \hat{P}_{b+1}(G_s) = P_{b}(G_s) \bigcup (x_{b+1}, y_{b+1}, z_{b+1}) \,,
\end{equation}
\aroundeqn
where $\hat{P}$ is the collection of 3D center for Gaussians $G_s$ of all currently visible regions, $(x_{b+1}, y_{b+1}, z_{b+1})$ is the Gaussians coordinate within the $b+1$ region.
We incorporate scenes from the subsequent bins into the previously constructed Gaussians with multiple surrounding frames as supervision. The incremental static Gaussian model $G_s$ is defined by:
\begin{equation}
\begin{aligned}
    \hat{C}(G_{s}) = \sum_{b=1}^{N} \Gamma_{b} \ \alpha_{b} \ C_{b}, \quad
     \Gamma_{b} = \prod_{i=1}^{b-1} (1-\alpha_{b}) \,,
\end{aligned}
\end{equation}
where $C$ denotes the color corresponding to each Gaussian in a certain view, $\alpha$ is the opacity, and $\Gamma$ is the accumulated transmittance of the scene according to $\alpha$ in all bins. During this process, the overlapping regions between surrounding multi-camera images are used to form the Gaussian models' implicit alignment jointly.

Note that during the incremental construction of static Gaussian models, there may be differences in sampling the same scene between the front and rear cameras. As such, we use a weighted averaging to reconstruct the scene's colors as accurately as possible during the 3D Gaussian projection:
\aroundeqn
\begin{equation}
  \tilde{C} = \varsigma(G_{s}) \sum \omega(\hat{C}(G_{s})|\mathbf{R,T}) \,,
  \label{eq:color}
\end{equation}
\aroundeqn
where $\tilde{C}$ is the optimized pixel color, $\varsigma$ denotes the differential splatting,  
$\omega$ is the weight for different views, $[\mathbf{R},\mathbf{T}]$ is the view-matirx for aligning multi-camera views.

\subsubsection{Composite Dynamic Gaussian Graph}
\begin{figure*}[t]
  \centering
  \includegraphics[width=0.9\linewidth]{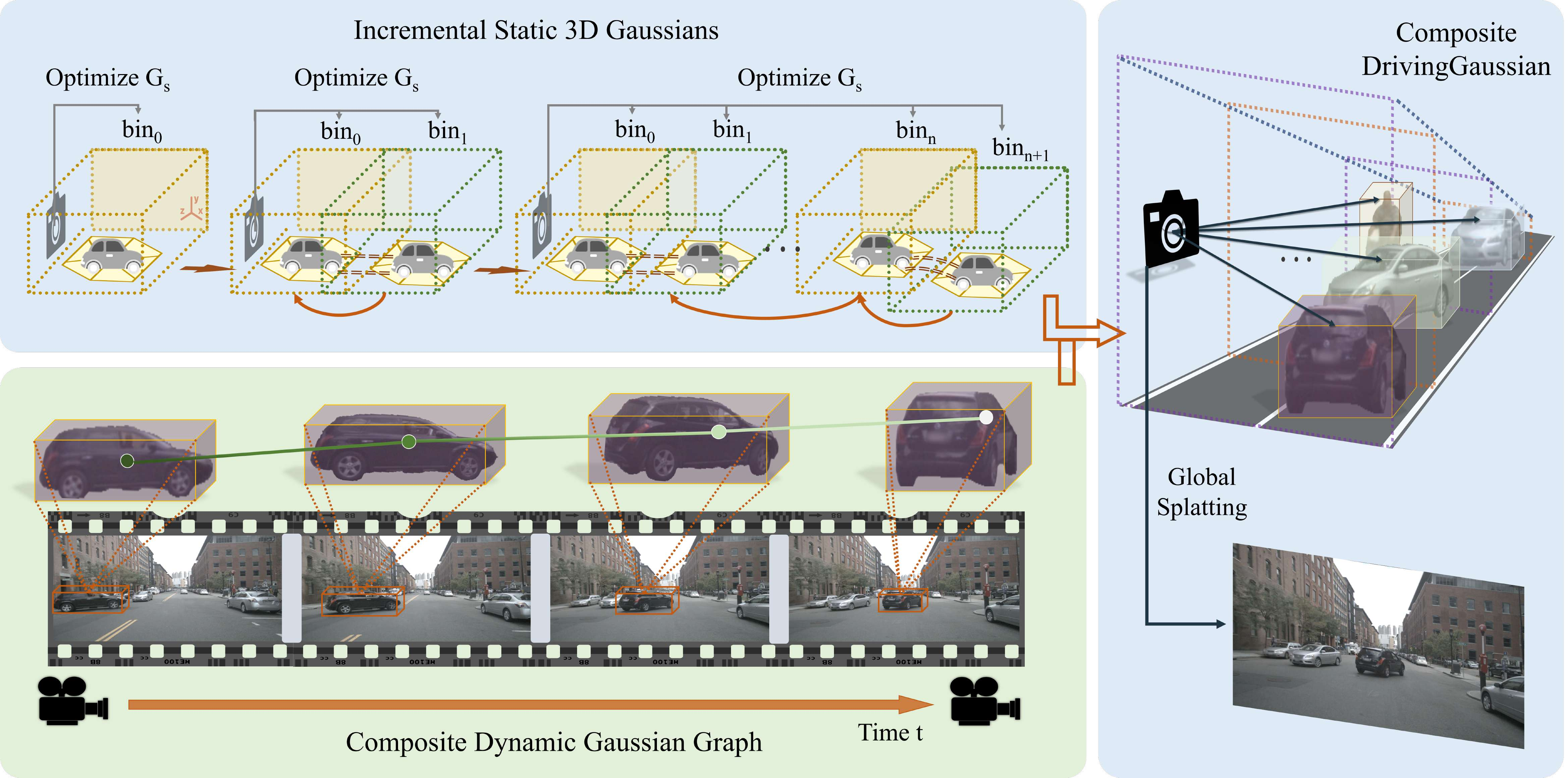}
  \caption{\textbf{Composite Gaussian Splatting with Incremental Static 3D Gaussians and Dynamic Gaussian Graph.} We adopt Composite Gaussian Splatting to decompose the whole scene into static background and dynamic foreground objects, reconstructing each part separately and integrating them for global rendering.}
  \label{static-dynamic}
  \afterfig
 \end{figure*}

The autonomous driving environment is highly complex, involving multiple dynamic objects and temporal changes. As shown in Figure~\ref{static-dynamic}, objects are often observed from limited views (e.g., 2-4 views) due to the egocentric movements of the vehicle and dynamic objects. In addition, fast moving objects also lead to significant appearance, making it challenging to represent them using fixed Gaussians.

To address the challenges, we introduce the Composite Dynamic Gaussian Graph, enabling the construction of multiple dynamic objects in long-term, large-scale driving scenes. 
We first decompose dynamic foreground objects from static backgrounds to build the dynamic Gaussian graph using bounding boxes provided by the datasets. Dynamic objects are identified by their object ID and the corresponding timestamps of appearance. Additionally, the Grounded SAM Models~\cite{ren2024grounded} are employed for precise pixel-wise extraction of dynamic objects based on the range of bounding boxes.

We construct a dynamic Gaussian graph using 
\aroundeqn
\begin{equation} 
H = <O, G_d, M, P, A, T> \,,
\end{equation} 
\aroundeqn
where each node stores an instance object $o \in O$, $g_i \in G_d$ denotes the corresponding dynamic Gaussians, and $m_o \in M$ is the transform matrix for each object. Here, $ p_{o}(x_t, y_t, z_t) \in P$ is the center coordinate of the bounding box, and $a_o = (\theta_{t}, \phi_{t}) \in A $ is the orientation of the bounding box at time step $t \in T$.
We compute one Gaussian separately for each dynamic object. Using the transformation matrix $m_o$, we transform the coordinate system of the target object $o$ to the world coordinate where the static background resides:
\aroundeqn
\begin{equation}
  \mathbf{m_{o}^{-1}} = \mathbf{R}_{o}^{-1}\mathbf{S}_{o}^{-1} \,,
  \label{eq:object}
\end{equation}
\aroundeqn
where $\mathbf{R}_{o}^{-1}$ and $\mathbf{S}_{o}^{-1}$ are the rotation and translation matrices corresponding to each object. 

After optimizing all nodes in the dynamic Gaussian graph, we combine dynamic objects and static backgrounds using a Composite Gaussian Graph. Each node's Gaussian distribution is concatenated into the static Gaussian field based on the bounding box position and orientation in chronological order. In cases of occlusion between multiple dynamic objects, we adjust the opacity based on the distance from the camera center: closer objects have higher opacity, following the principles of light propagation:
\aroundeqn
\begin{equation}
  \alpha_{o, t} = \sum \frac{(p_{t}-b_{o})^{2} \cdot \cot a_o}{\| (b_o | \mathbf{R_o,S_o}) - \rho \|^{2}} \alpha_{p_0} \,,
  \label{eq:opacity}
\end{equation}
\aroundeqn
where $\alpha_{o, t}$ is the adjusted opacity of Gaussians for object $o$ at time step $t$, $p_{t} = (x_t, y_t, z_t)$ is the center of Gaussians for the object. $ [\mathbf{R_o, S_o}] $ denotes the object-to-world transform matrix, $\rho$ denotes the center of camera view, and $\alpha_{p_{0}}$ is the opacity of Gaussians.

The composite Gaussian field, including both static background and multiple dynamic objects, is formulated by:
\aroundeqn
\begin{equation}
  G_{comp} = \sum H <O, G_d, M, P, A, T> + G_{s} \,,
  \label{eq:composite}
\end{equation}
\aroundeqn
where $G_s$ is obtained in Section~\ref{static} through Incremental Static 3D Gaussians and $H$ denots the optimized dynamic Gaussian graph.

3D driving scene editing is based on the composite Gaussians of static background and dynamic objects, reconstructed by Composite Gaussian Splatting, and performs multiple editing tasks on them without extra training.

\subsection{Global Rendering via Gaussian Splatting} 
\label{rendering}

We adopt the differentiable 3D Gaussian splatting renderer $\varsigma$ from~\cite{kerbl20233d} and project the global composite 3D Gaussian into 2D with the covariance matrix $\widetilde{\Sigma}$:
\aroundeqn
\begin{equation}
  \widetilde{\mathbf{\Sigma}} = \mathbf{J} \mathbf{E} \ \mathbf{\Sigma} \ \mathbf{E^\top} \mathbf{J^\top} \,,
  \label{eq:cocariance}
\end{equation}
\aroundeqn
where $\mathbf{J}$ is the Jacobian matrix of the perspective projection, and $\mathbf{E}$ denotes the world-to-camera matrix.

The composite Gaussian field projects the global 3D Gaussian onto multiple 2D planes and is supervised using surrounding views at each time step. In the global rendering process, Gaussians from the next time step are initially invisible to the current and subsequently incorporated with the supervision of the corresponding global images.

The loss function of our method consists of three parts. Similar to~\cite{kerbl20233d, xie2023s3im}, we introduce the Tile Structural Similarity (TSSIM) to Gaussian Splatting, which measures the similarity between the rendered tile and the corresponding ground truth:
\aroundeqn
\begin{equation}
  L_{TSSIM}(\delta) = 1 - \frac{1}{Z}\sum_{z=1}^{Z} SSIM(\Psi(\hat{C}), \Psi(C)) \,,
  \label{eq:TSSIM loss}
\end{equation}
\aroundeqn
where we split the screen into $M$ tiles, $\delta$ is the training parameters of the Gaussians, $\Psi(\hat{C})$ denotes the rendered tile from Composite Gaussian Splatting, and $\Psi(C)$ denotes the paired ground-truth tile.

We also introduce a robust loss to reduce outliers in 3D Gaussians, which can be defined as:
\aroundeqn
\begin{equation}
  L_{Robust}(\delta) = \kappa(\|\hat{I} - I\|_{2}) \,,
  \label{eq:robust loss}
\end{equation}
\aroundeqn
where $\kappa \in (0,1]$ is the shape parameter that controls the robustness of the loss, $I$ and $\hat{I}$ denote the ground truth and the synthesis image, respectively.

The LiDAR loss is further employed by supervising the expected Gaussians' position from the LiDAR, obtaining better geometric structure and edge shapes:
\aroundeqn
\begin{equation}
  L_{LiDAR}(\delta) = \frac{1}{s}\sum \|P(G_{comp}) - L_s\|^{2} \,,
  \label{eq:LiDAR loss}
\end{equation}
\aroundeqn
where $P(G_{comp})$ is the position of 3D Gaussians, and $L_s$ is the LiDAR point prior. 

We optimize the Composite Gaussians by minimizing the sum of three losses in Eq~\ref{eq:TSSIM loss}-\ref{eq:LiDAR loss}.
The proposed editing method leverages globally rendered images to identify editing targets and utilizes depth information derived from 3DGS as geometry priors, enabling effective and realistic multitask editing.

\subsection{Controllable Editing for Dynamic Driving
Scenes}
We tackle three key editing tasks for autonomous driving simulations: texture modification, weather simulation, and object manipulation. To support these diverse editing tasks, we have developed a framework that sequentially operates on the Gaussians of the reconstructed scene using 3D geometric priors, Large Language Models (LLMs) for dynamic predictions, and advanced editing techniques to ensure overall coherence and realism.

\noindent \textbf{Texture Modification:} This task involves applying patterns to the surfaces of 3D objects. In autonomous driving, texture modification extends beyond aesthetics to allow the addition of critical road features, such as cracks, manhole covers, and signage, which are crucial to building more robust testing environments.
We show failure cases of object detection model~\cite{liu2023grounding} in Figure~\ref{fig:perception}, highlighting the importance of editing simulation. 
Before editing, the perception model accurately identifies objects within the scene. However, after editing with \ourmethod{}, challenging cases in the 3D scene become undetectable to the model, providing a more effective testing environment to assess the reliability and robustness of various components within autonomous driving systems.

\textbf{Weather Simulation:} This task focuses on integrating dynamic meteorological phenomena, such as precipitation, snowfall, and fog, into autonomous driving scenarios. Weather simulation is critical for replicating driving conditions in severe weather, demonstrating its importance in augmenting training datasets.

\textbf{Object Manipulation:} This task is divided into object deletion and insertion within the reconstructed scene. Object insertion is further categorized into static and dynamic types, with dynamic insertion adaptively predicting the object's motion trajectory. These manipulations are essential for building robust autonomous driving simulation systems.

\begin{figure*}[tp]
  \centering
  \includegraphics[width=0.9\linewidth]{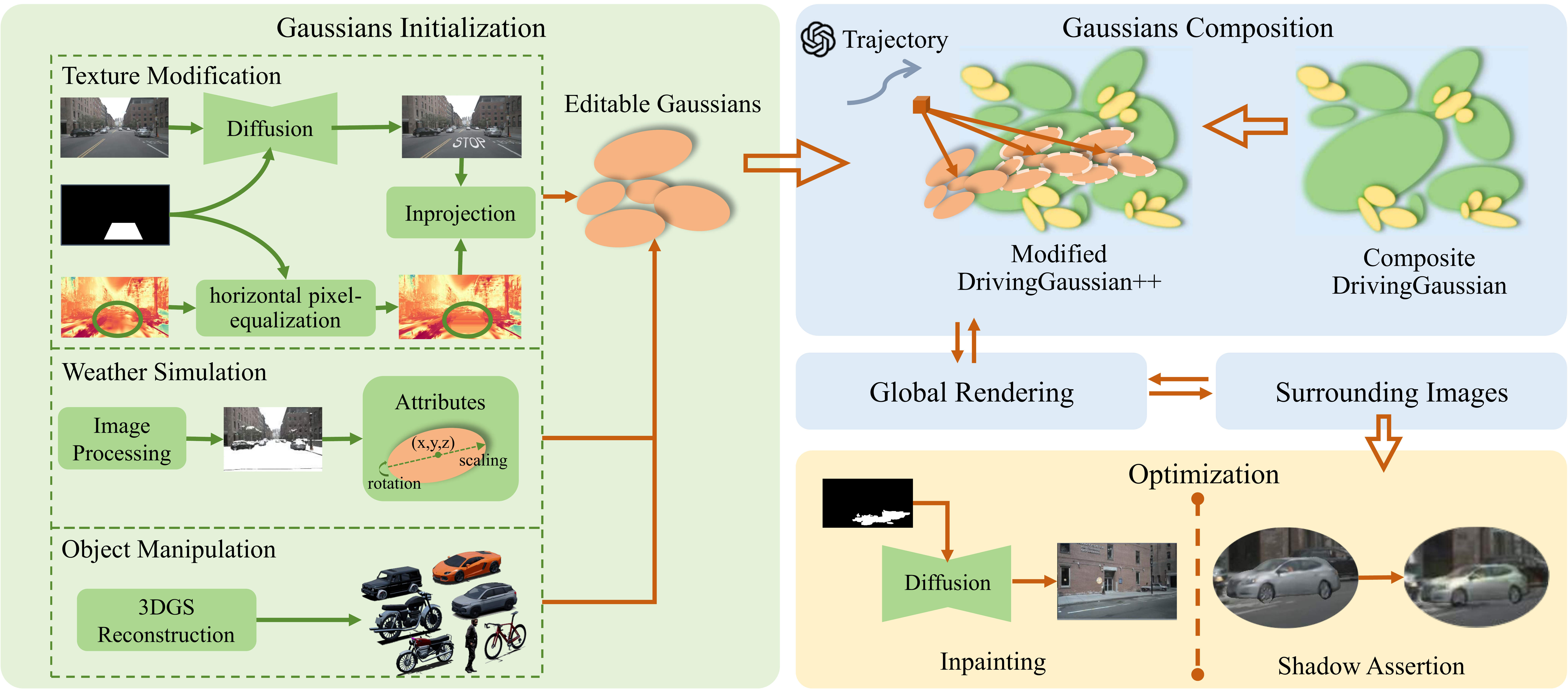}
  \caption{\textbf{Pipeline of our editing framework.} \textbf{Left:} We separately determine the target Gaussians for diverse tasks. For texture modification, first edit images and then conduct inverse projection with depth information. For weather simulation, design the attributes and distribution of weather particles in detail. For object manipulation, use models from the foreground bank as inserted objects, or delete objects based on annotations. \textbf{Top right:} We composite the Gaussians and predict the trajectory of new objects with LLM. \textbf{Bottom right:} To produce realistic editing, we perform shadow addition and inpainting. Our method achieves training-free, multi-task editing specifically for dynamic driving scenarios. 
  }
  \label{fig:edit_overview}
  \afterfig
\end{figure*}

\begin{figure}[!t]
  \centering
  \includegraphics[width=0.9\linewidth]{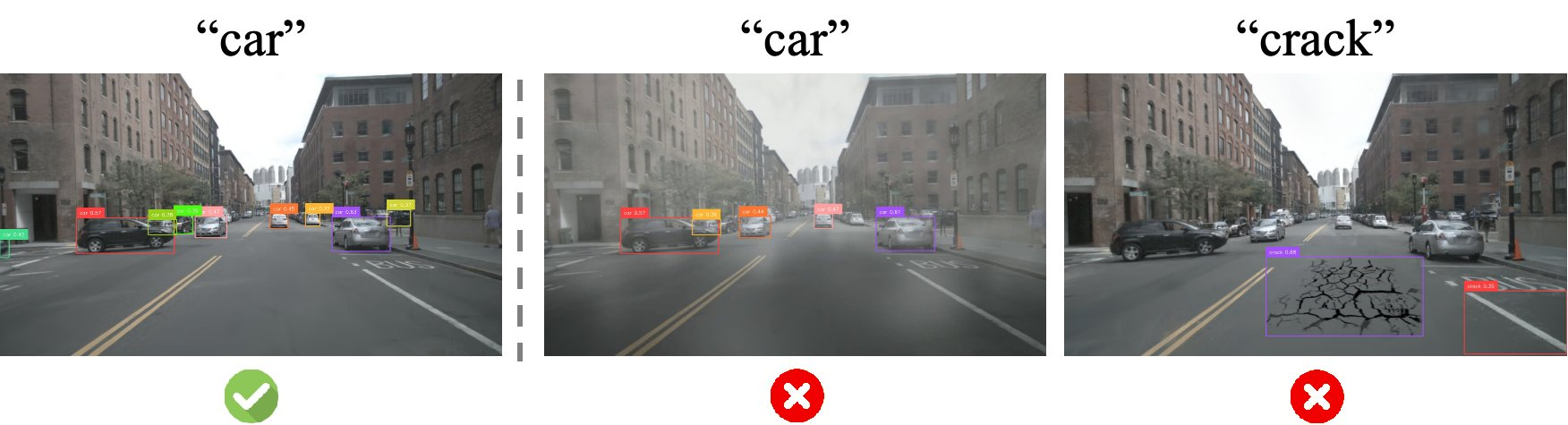}
  \caption{\textbf{Failed Object Detection Cases Simulation.} We use GroundingDINO~\cite{liu2023grounding} as the object detection model. It struggles to completely identify vehicles in foggy weather and fails to accurately recognize textures, such as cracks.}
  \label{fig:perception}
  \vspace{-10pt}
\end{figure}

To enable multitask editing, we propose a framework that performs sequential operations on the Gaussians of the reconstructed scene without extra training. 
The process begins by identifying target Gaussians to edit using 3D geometric priors, followed by their integration into the scene. We employ large-language models (LLMs) to predict dynamic object trajectories and apply image-processing techniques to refine the results, ensuring coherence and realism. The editing pipeline is illustrated in Figure~\ref{fig:edit_overview}.

\subsubsection{Initialization}\label{3.1:creation}
In the proposed editing framework, we refer to those introduced to or removed from the original scene as target Gaussians, while those reconstructed from the initial scene are termed original Gaussians. 
The approach for determining target Gaussians depends on the specific editing task. For object removal, the target Gaussians, corresponding to the subset of original Gaussians marked for removal, are identified by refining the 3D bounding box provided by the dataset. 
Since the LiDAR prior is integrated during the reconstruction process, we can accurately locate their positions without additional alignment of the coordinate system. For other editing tasks, new Gaussians are generated as target Gaussians, designed with specific shapes and distributions to meet the requirements of each task.

\noindent \textbf{Texture Modification.} We enhance the surface texture of an object by introducing new flat Gaussians onto the surface of the designated editing region. The process begins by selecting a viewpoint and using a diffusion model or similar tools to edit the original image, generating a target image and a corresponding mask to guide the 3D editing. Specifically, we randomly select a viewpoint that provides clear visibility of the target region and render the image to be edited along with its associated depth map. Next, we define the 2D mask of the target region and apply the diffusion model or image processing software to modify the image in a 2D space, producing the target image. 

\begin{figure}[t]
  \centering
  \includegraphics[width=\linewidth]{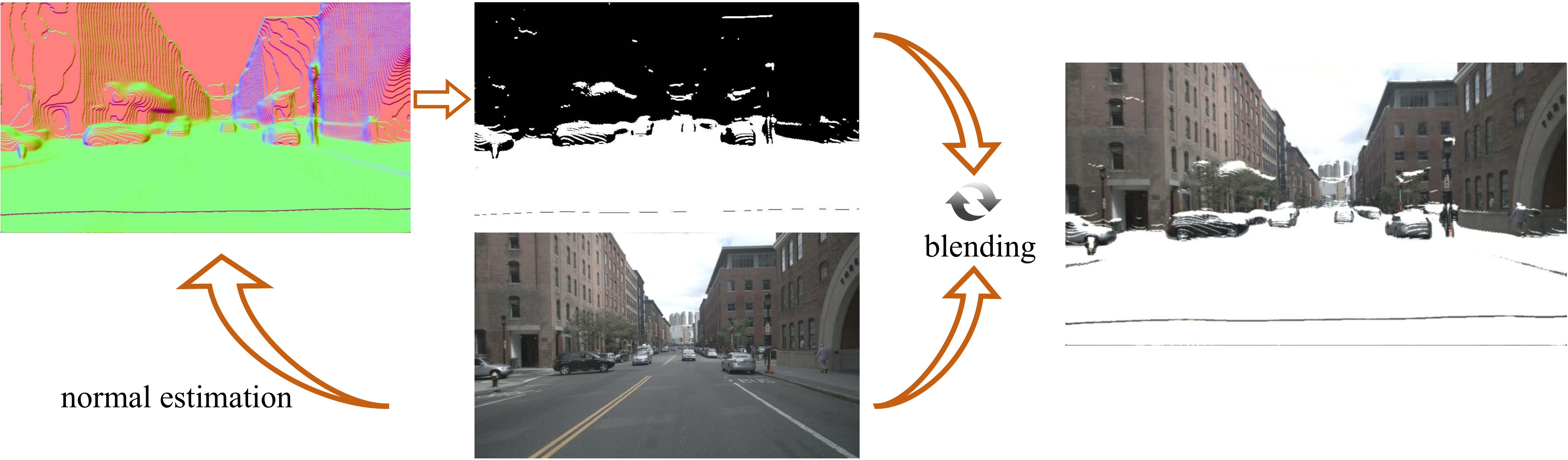}
  \caption{\textbf{Implementation of Snow Coverage Effect.} We compute image normals, generate a snow mask, and fuse it with the original image to produce the target image. Depth priors are used to back-project the result into 3D space.}
  \label{fig:fig_snow}
  \afterfig
 \end{figure}

Using the target image and the mask, we generate target Gaussians and assign appropriate attributes through inverse projection. As shown in Fig.~\ref{fig:fig_snow}, \ourmethod{} projects the edited content onto the corresponding position based on the rendered depth map and pixel-wise correspondence.

However, discrepancies may arise between the surface reconstructed by 3D Gaussian Splatting and the actual object's surface. These discrepancies can lead to inconsistencies between the rendered depth and the real depth of the object, potentially causing the surface of the target Gaussians to appear uneven and unrealistic, thereby compromising the editing quality.

To address this issue, we perform equalization on the depth map. Specifically, we normalize the depth of the editing area to ensure a relatively uniform depth distribution along the horizontal axis, while preserving the depth distribution along the vertical axis: 

\aroundeqn
\begin{equation} 
    D_{opt}(M_{edit},x,y) = Average(D_{ori}(M_{edit},y)) \,,
\label{eq:horizontal}  
\end{equation} 
\aroundeqn
where $D_{ori}$, $D_{opt}$ separately represent the rendered depth before and after the depth equalization, $M_{edit}$ denotes the binary mask of the editing region and $x$, $y$ are the image coordinates. This approach yields a flat surface for the target Gaussians, significantly enhancing the visual quality and realism of the texture modification.

\noindent \textbf{Weather Simulation.} We simulate weather particles by incorporating Gaussians with specific physical properties into the current scene and achieve dynamic effects by adjusting the positions of these Gaussians at each time step. The first step in weather simulation is to design particles that align with the desired physical properties. We calculate the number of original Gaussians along with the range of their positions and introduce new Gaussians with specific shapes and colors in a particular distribution within the scene. Specifically, we use narrow, semi-transparent, white Gaussians to represent raindrops, irregular white ellipsoidal Gaussians to represent snowflakes, and Gaussians following a random distribution in the scene to represent fog. As an example, for snow simulation, we define the target Gaussians \( G_{snow} \) by:
\aroundeqn
\begin{equation}
    G_{snow} = \{G_k\}\,,  
\label{eq:cli}  
\end{equation} 
\aroundeqn
where the $k_{th}$ Gaussian $G_k$ satisfies $p_k = \zeta * (p_{\max} - p_{\min}) + p_{\min}$, $c_k = (1,1,1) + \epsilon$, $s_{k,y} = \min(\min(s_{k,x}, s_{k,x}), 0) + \varepsilon$, and $p_k$, $c_k$, $s_{k,y}$ separately denotes its 3D coordinate, color, and scale attribute. 
 
Second, to realize dynamic weather effects that include the falling of raindrops, the drifting of snowflakes, and the spread of fog, we add a specific trajectory to weather Gaussians according to the current time step. We describe the trajectory of snowflakes using an example:
\aroundeqn
\begin{equation} 
\begin{aligned}
    p_{k,t+1} &= p_{k,t} + trajj\_func(t) \,,
\end{aligned}
\label{eq:clitrajj}  
\end{equation} 
\aroundeqn
where $p_{k,t}$ denotes the position of $k_{th}$ Gaussian in $G_{snow}$ at timestep $t$, and $trajj\_func$ is a function that calculates the relative movement between consecutive positions in the time sequence. 

 \begin{figure*}[ht]
  \centering
  \includegraphics[width=.9\linewidth]{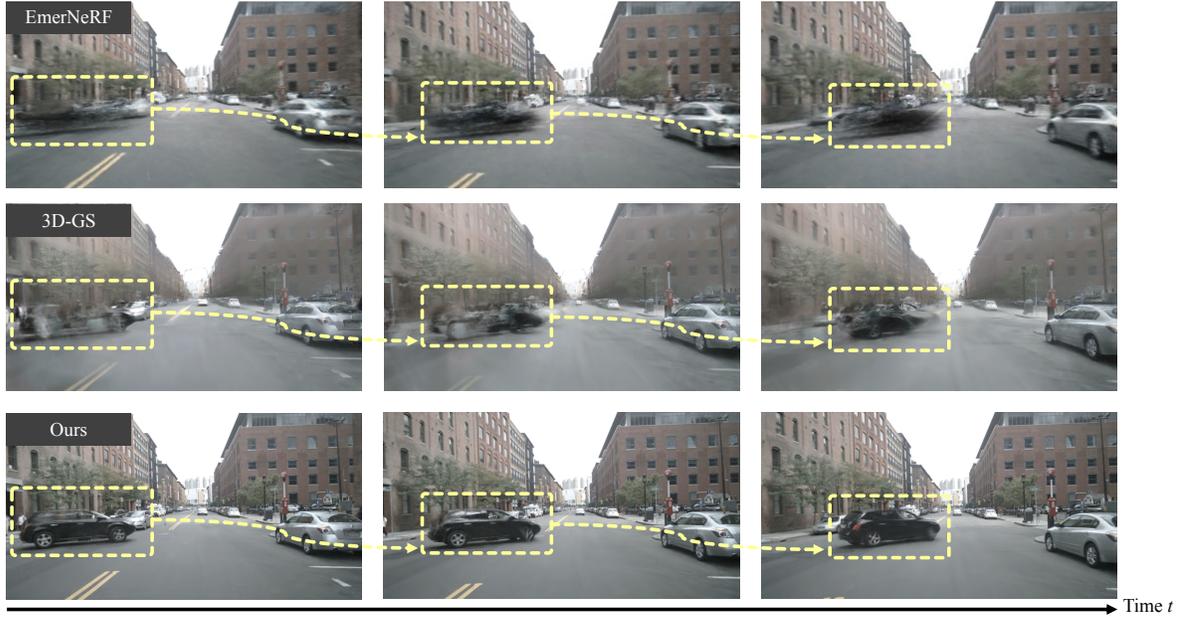}
  \caption{\textbf{Qualitative comparison} with EmerNeRF~\cite{yang2023emernerf} and 3DGS~\cite{kerbl20233d} on dynamic reconstruction for 4D driving scenes of nuScenes. \ourmethod{} enables the high-quality reconstruction of dynamic objects at high speed while maintaining temporal consistency.}
  \label{temporal_compare}
  \afterfig
 \end{figure*}

We also implement the 3D snow coverage effect, as shown in Figure~\ref{fig:fig_snow}. Specifically, we first calculate the normal map of training images based on Depth-Anything~\cite{yang2024depth} and the Sobel filter~\cite{kanopoulos1988design} as:
\aroundeqn
\begin{equation} 
    s_{i,x}, s_{i,y}= Sobel_x(D_i), Sobel_y(D_i) \,,
\end{equation} 
\begin{equation} 
    N_i = (\frac{s_{i,x}}{\sqrt{s_{i,x}^2+s_{i,y}^2}}, \frac{s_{i,y}}{\sqrt{s_{i,x}^2+s_{i,y}^2}}, \frac{1}{\sqrt{s_{i,x}^2+s_{i,y}^2}}) \,,
\label{eq:normal}  
\end{equation} 
\aroundeqn
where $D_i$ and $N_i$ denote the depth and normal map of the image $i$, while $s_{i,x}$ and $s_{i,y}$ are gradient magnitudes in the horizontal and vertical directions.
Based on the normal map, a snow mask is added in the region with a large vertical $y$ component. Using a processed image with snow coverage and rendered depth, an inverse projection is taken to calculate the snow particle Gaussians' 3D positions from this viewpoint. Finally, we combine the positions under different viewpoints to realize a consistent snow-covering effect between frames. To avoid inconsistencies arising from repeated calculations in overlapping areas between frames, we construct a KD-Tree, and prune nodes that are closer to each other:
\aroundeqn
\begin{equation} 
\begin{aligned}
    P_{snow} = \bigcup_{i=1}^n{[P_{i} - KNN(\bigcup_{j=1}^{i-1}{P_{j}},\ KDT(P_{i}),\ \text{1}_{th})]}\,,
\end{aligned}
\label{eq:knn}  
\end{equation} 
\aroundeqn
where $P_{snow}$ denotes positions of the target snow particle Gaussians, $P_i$ denotes positions calculated from viewpoints in the $i_{th}$ frame, $KDT$ refers to the constructed KD-Tree, and $KNN$ stands for K Nearest Neighbor (KNN) function, which takes three parameters as input: \textit{the search range}, \textit{the KDTree of the search target}, and \textit{the number of top k neighbors}. We insert the final target snow particle Gaussians into the scene and realize the snow-covering effect.

\noindent \textbf{Object Manipulation.} Due to the distinct nature of the operations, object insertion and deletion differ in their implementation. For object removal, the target Gaussians correspond to the object to be deleted. First, we extract the object's 3D bounding box matrix from the dataset annotations and crop the Gaussians within the bounding box. To address holes caused by insufficient reconstruction in the occluded areas, we further use a diffusion model to locally paint the rendered image (Section~\ref{3.4:optimization}). For object insertion, we construct a 3D foreground bank containing objects reconstructed using 3D Gaussian Splatting, which can be directly utilized for insertion. The objects in the bank are acquired through 3DGS reconstruction of Blender models collected online and sparse reconstruction of vehicles from autonomous driving datasets. Additionally, the lighting of the foreground object can be adjusted using MCLight~\cite{wei2024editable} to better match the current scene.

\subsubsection{Gaussians Composition with Trajectory Prediction}\label{3.2:trajj_pred}
After identifying the target Gaussians, we integrate them with the original scene. This process aligns both components within the same coordinate system to establish physically accurate occlusion relationships. Notably, the covariance matrices of the two Gaussian groups may interfere with each other during rasterization rendering, potentially leading to blurry results. Thus, we perform an additional forward process for the added Gaussians and store the covariance matrices of the transformed Gaussians. Finally, the combined scene is rendered for visualization.
 \begin{figure}[tp]
  \centering
  \includegraphics[width=\linewidth]{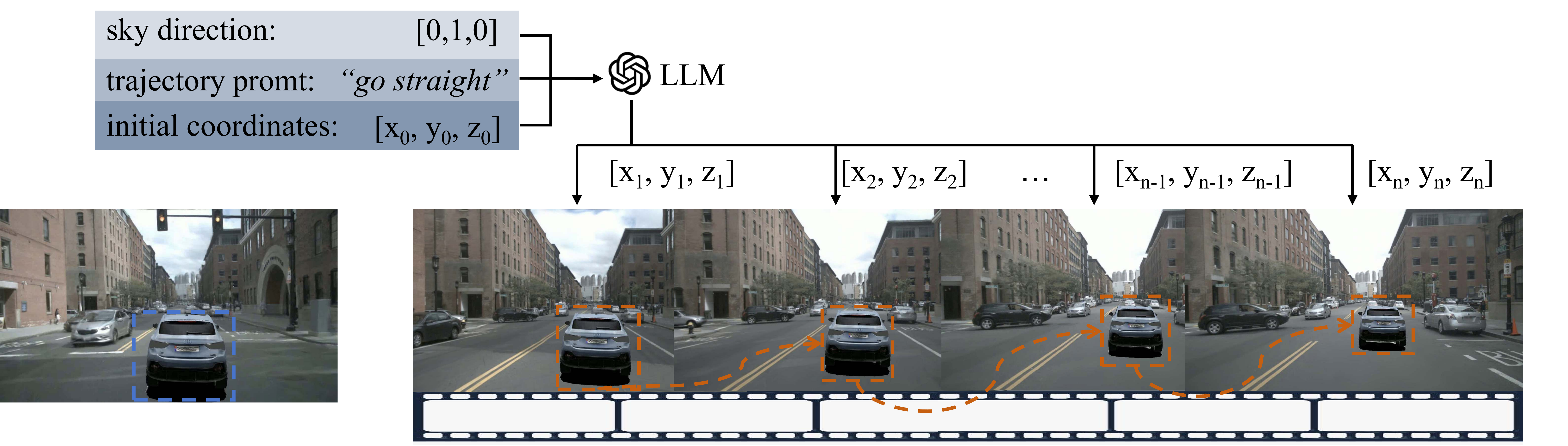}
  \caption{\textbf{Trajectory Prediction with LLM.} With the sky direction, trajectory prompt, and initial position, we utilize LLM to predict the future trajectory of the added car.}
  \label{fig:fig_llm}
  \afterfig
 \end{figure}

For the object insertion task, to ensure that dynamic objects have reasonable and diverse motion trajectories, we utilize the powerful scene comprehension capability of the Large Language Models~\cite{achiam2023gpt} to predict the future trajectories of the inserted objects as:
\aroundeqn
\begin{equation} 
\begin{aligned}
   P_{t+1} &= P_{t} + Trajj\_pred(t) \,, \\
   Trajj\_pred &= LLM(P_0,\ dir_{sky},\ des) \,,
\end{aligned}
\label{eq:LLM}  
\end{equation} 
\aroundeqn
where $P_t$ denotes the position of inserted object at timestep $t$ while $P_0$ is the initial position, $Trajj\_pred(t)$ denotes the relative positions at time step $t$ generated by LLM, $dir_{sky}$ is the sky direction and $des$ denotes the description of the expected trajectory. Specifically, we take \textit{the initial vehicle's position}, \textit{the sky direction}, and \textit{the trajectory description} as prompts, and generate a series of possible future trajectory sequences through GPT-4o~\cite{achiam2023gpt}.

\subsubsection{Global Refinement with Differentiable Rendering}
\label{3.4:optimization}

Leveraging the recent advances in diffusion models and 2D image processing, our approach integrates these techniques to enhance the results of object manipulation tasks. For object removal, we use diffusion models to locally inpaint the damaged regions of the rendered image. First, we delete the target Gaussians of the specified region based on 3D annotations. 
However, due to occlusions and limitations in the data acquisition viewpoint, surrounding areas of the deleted Gaussians often contain artifacts or holes with poor reconstruction quality. 
To address this issue, we use the K Nearest Neighbor lgorithm to identify a set of Gaussians requiring repair around the target region. We then perform a binarization rendering on these Gaussians to generate the corresponding inpainting masks:
\aroundeqn
\begin{equation} 
\begin{aligned}
   M_{inpaint} = \{G_i \in G_{l}\ |\ d(p_i,\ G_{del}) < d_{thr}\} \,,
\end{aligned}
\label{eq:knn2}  
\end{equation} 
\aroundeqn
where $M_{inpaint}$ is a binary mask with the Gaussians to be inpainted set to 1, $G_{l}$ represents the remaining Gaussians after removal, while $G_{del}$ denotes the removed Gaussians, $p_i$ denotes the position of $G_i$, $d_{thr}$ is the distance threshold that determines which Gaussians should be inpainted. The nearest distance between $G_i$ and the Gaussians in $G_{del}$ is given by $d(p_i, G_{del})$, which is computed as $d(p_i, G_{del}) = KNN(p_i, KDT(P_{del})$. Subsequently, the images to be repaired, along with corresponding masks, are fed into the diffusion model as inputs. \ourmethod{} performs partial inpainting to restore the integrity and visual authenticity of the scene, achieving more realistic and seamless object removal.

 \begin{figure}[tp]
  \centering
  \includegraphics[width=\linewidth]{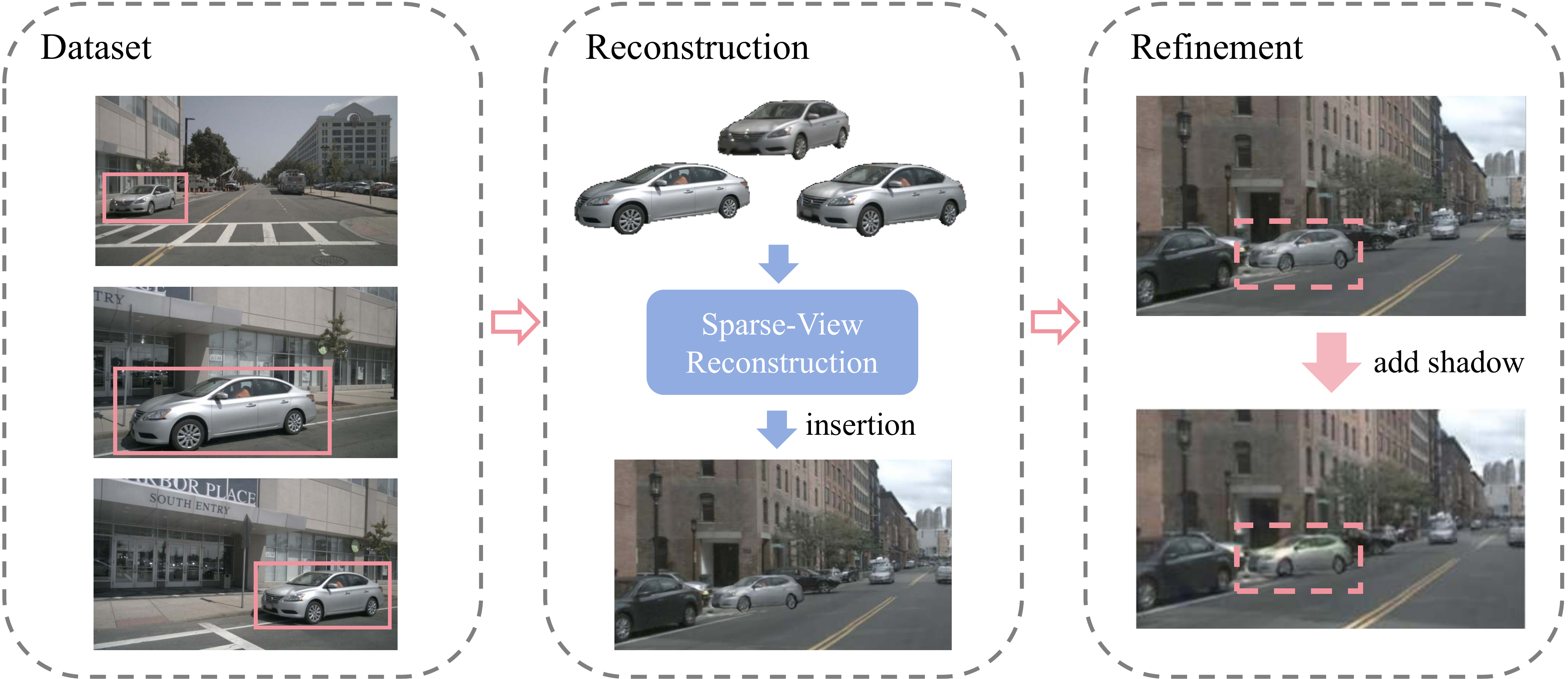}
  \caption{\textbf{Refinement on Object Insertion.} We sparsely reconstruct the car from nuScenes with one-shot generation and then generate the shadow for the car.}
  \label{fig:fig-opt}
  \afterfig
 \end{figure}
 
For the object insertion task, when extracting data from autonomous driving datasets, we perform sparse reconstruction to generate the target Gaussians. The reconstructed vehicles lack shadow information, which leads to a levitation effect in the rendered images. To enhance the realism of object insertion without additional training, we adopt a shadow synthesis approach inspired by ARShadowGAN~\cite{liu2020arshadowgan}. Specifically, we synthesize shadows for the inserted objects to eliminate the levitation effect, thereby improving the visual consistency and realism of the scene.

\section{Experiments}
\begin{table*}[!t]
  \caption{
  {\textbf{Overall perforamnce of \ourmethod{} with existing state-of-the-art approaches on the nuScenes dataset.} Ours-S denotes the \ourmethod{} with SfM initialization, and Ours-L denotes training with LiDAR prior.}}
  \centering
  \footnotesize
  \setlength{\tabcolsep}{8mm}{
    \begin{tabular}{ccccc}
 \toprule
    \textbf{Methods} & \textbf{Input} & \textbf{PSNR $\uparrow$} & \textbf{SSIM $\uparrow$} & \textbf{LPIPS $\downarrow$} \\
    \hline 
    Instant-NGP~\cite{muller2022instant}           & Images         & 16.78 & 0.519 & 0.570  \\
    NeRF+Time                              & Images         & 17.54 & 0.565 & 0.532  \\
    Mip-NeRF~\cite{barron2021mip}          & Images         & 18.08 & 0.572 & 0.551  \\
    Mip-NeRF360~\cite{barron2022mip}       & Images         & 22.61 & 0.688 & 0.395  \\
    Urban-NeRF~\cite{rematas2022urban}     & Images + LiDAR & 20.75 & 0.627 & 0.480  \\
    S-NeRF~\cite{xie2023s}                 & Images + LiDAR & 25.43 & 0.730 & 0.302  \\
    SUDS~\cite{turki2023suds}              & Images + LiDAR & 21.26 & 0.603 & 0.466  \\
    EmerNeRF~\cite{yang2023emernerf}       & Images + LiDAR &  26.75 &  0.760 & 0.311  \\
    \hline
    3DGS~\cite{kerbl20233d}               & Images + SfM Points & 26.08 & 0.717 &  0.298  \\
    \hline
    Ours-S                                 & Images + SfM Points & \cellcolor{tabsecond} 28.36 & \cellcolor{tabsecond} 0.851 & \cellcolor{tabsecond} 0.256  \\
    \textbf{Ours-L}                        & Images + LiDAR      & \cellcolor{tabfirst} 28.74  & \cellcolor{tabfirst} 0.865 & \cellcolor{tabfirst} 0.237  \\
    \hline
    \end{tabular}%
    }
  \label{compare_SOTA}%
  \aftertab
\end{table*}%

\subsection{Datasets}

The nuScenes~\cite{caesar2020nuscenes} dataset comprises 1000 driving scenes collected using multiple sensors (6 cameras, 1 LiDAR, etc.). The images are annotated with accurate 3D bounding boxes from 23 object classes. 
Our experiments utilize the keyframes of six challenging scenes with surrounding views, collected from 6 cameras and corresponding LiDAR sweeps (optional), as input.
The KITTI-360~\cite{liao2022kitti} dataset contains multiple sensors, corresponding to over 320k images and point clouds. Although the dataset provides stereo camera images, we use only a single camera to demonstrate that our method also performs well in monocular scenes.

\subsection{Implementation Details}
Our implementation is based on the 3DGS framework, with fine-tuned optimization parameters to fit the large-scale unbounded scenes. Instead of using SfM points or randomly initialized points as input, we employ the LiDAR prior mentioned in Section~\ref{LiDAR} for initializations. Considering the computational cost, we use a voxel grid filter for LiDAR points, reducing the scale without losing geometric features. We employ random initialization for dynamic objects with initial points set to 3000, since objects are relatively small in large-scale scenes. We increase the total training iterations to 50,000, set the threshold for densifying grad to 0.001, and reset the opacity interval to 900. The learning rate of Incremental Static 3D Gaussians remains the same as in the official setting, while the learning rate of the Composite Dynamic Gaussian Graph exponentially decays from 1.6e-3 to 1.6e-6. We assess our models using various metrics, including PSNR, SSIM, and LPIPS, and report the average results of all camera frames in the scenes. All experiments are carried out on an 8 RTX8000 with 384 GB memory.

\begin{table*}[ht]
\footnotesize
    \renewcommand\arraystretch{1.5}
    \setlength{\tabcolsep}{0.005\linewidth}
    \centering
    \caption{\textbf{Editing Efficiency.} We demonstrate the efficient multi-task editing capabilities of our method in comparison with existing state-of-the-art 3D editing approaches. The evaluated multi-task editing scenarios include object manipulation, weather simulation, texture modification, and dynamic editing. ``Exec. Time'' indicates the total execution time for completing each editing task under a fair comparison setting. 
    }
    \label{tab:methods_comparison}
    \begin{tabular}{lcccccccccc}
    \toprule
    \textbf{Method} & \textbf{Object Manipulation} & \textbf{Weather Simulation} & \textbf{Texture Modification} & \textbf{Dynamic Editing} & \textbf{Exec. Time} & $\textbf{CLIP}_{dir} \uparrow$\\
        \hline

InstructNeRF2NeRF & \textcolor{red}{\usym{2717}} & \textcolor{ForestGreen}{\usym{2713}} & \textcolor{ForestGreen}{\usym{2713}} & \textcolor{red}{\usym{2717}} & $\approx$274 minutes & 0.1570 \\
%
InstructGS2GS & \textcolor{red}{\usym{2717}} & \textcolor{ForestGreen}{\usym{2713}} & \textcolor{ForestGreen}{\usym{2713}} & \textcolor{red}{\usym{2717}} & $\approx$60 minutes & 0.0918 \\
ClimateNeRF & \textcolor{red}{\usym{2717}} & \textcolor{ForestGreen}{\usym{2713}} & \textcolor{red}{\usym{2717}} & \textcolor{red}{\usym{2717}} & $\approx$107 minutes & 0.1105 \\
%
%
\midrule
\textbf{Ours} & \textcolor{ForestGreen}{\usym{2713}} & \textcolor{ForestGreen}{\usym{2713}} & \textcolor{ForestGreen}{\usym{2713}} & \textcolor{ForestGreen}{\usym{2713}} & \textbf{$\approx$8 minutes} & \textbf{0.2327}\\
\bottomrule     
    \end{tabular}
    \label{table:efficiency}
    \vspace{-2mm}
\end{table*}

\begin{figure*}[!t]
  \centering
  \includegraphics[width=.9\linewidth]{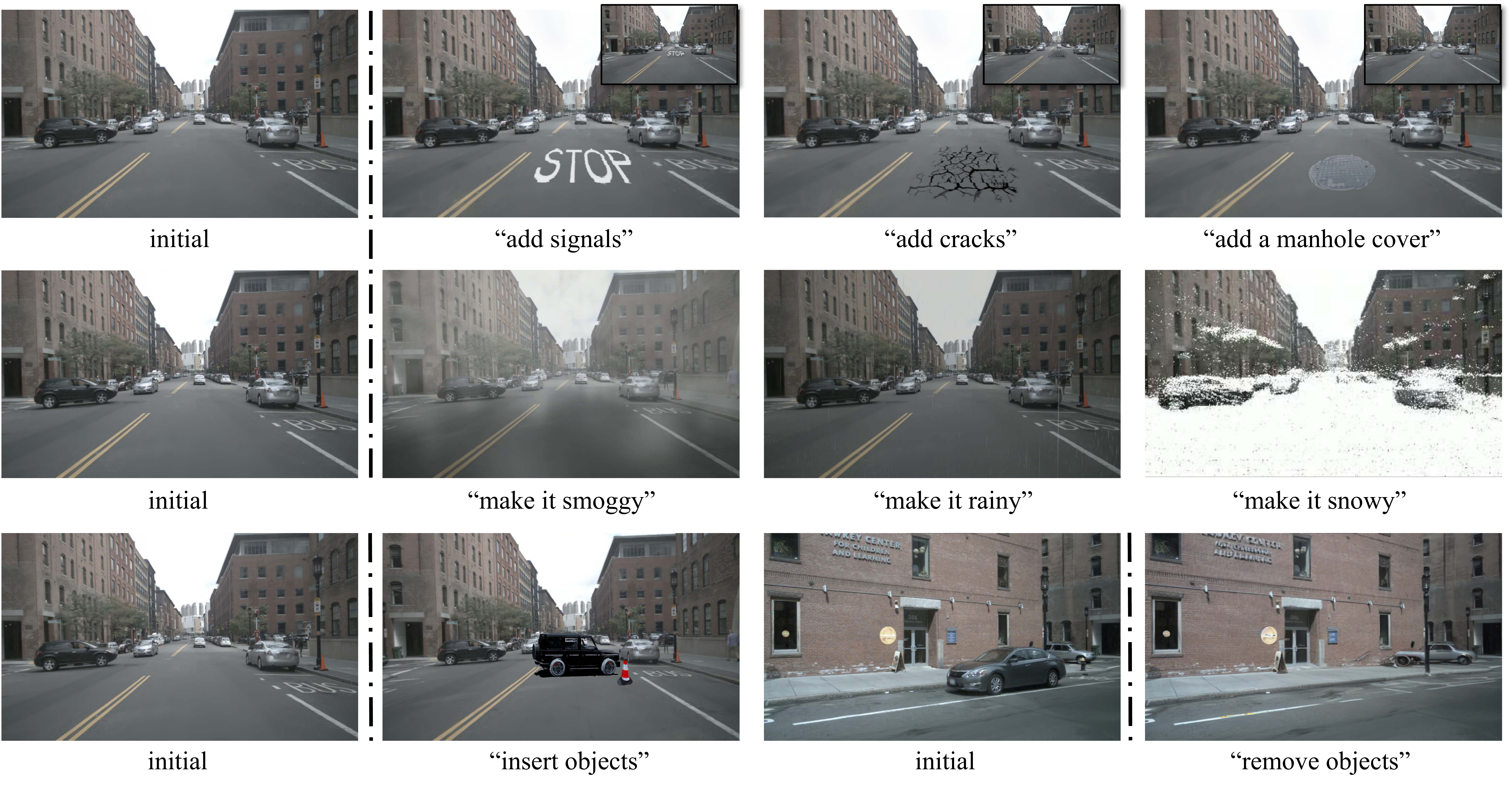}
  \caption{\textbf{Editing Results of nuScenes dataset.} We demonstrate the results of \ourmethod{} across different tasks. \ourmethod{} enables realistic and consistent 3D editing of texture, weather, and objects in driving scenes.}
  \label{fig:res}
  \vspace{-8pt}
\end{figure*}

\begin{table}[!t]
  \caption{
  {\textbf{Overall perforamcne on the KITTI-360.} Comparisions of \ourmethod{} with existing state-of-the-art approaches on the KITTI-360 dataset.}}
  \centering
  \footnotesize
  \setlength{\tabcolsep}{5.5mm}{
    \begin{tabular}{ccc}
    \toprule
    \textbf{Methods} & \textbf{PSNR $\uparrow$} & \textbf{SSIM $\uparrow$} \\
    \hline 
    NeRF~\cite{mildenhall2021nerf}        & 21.94 & 0.781  \\
    Point-NeRF~\cite{xu2022point}         & 21.54 & 0.793   \\
    NSG~\cite{ost2021neural}              & 22.89 & 0.836   \\
    Mip-NeRF360~\cite{barron2022mip}      & 23.27 & 0.836   \\
    SUDS~\cite{turki2023suds}             & 23.30 & 0.844   \\
    DNMP~\cite{lu2023urban}               &  23.41 & 0.846   \\
    \hline
    Ours-S                                & \cellcolor{tabsecond} 25.18 & \cellcolor{tabsecond} 0.862   \\
    \textbf{Ours-L}                       & \cellcolor{tabfirst} 25.62  & \cellcolor{tabfirst} 0.868  \\
    \hline
    \end{tabular}%
    }
  \label{KITTI-360}%
  \aftertab
  \vspace{-2mm}
\end{table}%

\subsection{Reconstruction Results and Comparisons}

\subsubsection{Comparisons of surrounding views on nuScenes.}
We evaluate the proposed model against state-of-the-art approaches, including NeRF-based methods~\cite{muller2022instant, barron2021mip, barron2022mip, rematas2022urban, xie2023s, turki2023suds, yang2023emernerf} and 3DGS-based schemes~\cite{kerbl20233d}.
As shown in Table~\ref{compare_SOTA}, our method outperforms Instant-NGP~\cite{muller2022instant}, which employs a hash-based NeRF for novel view synthesis.
While Mip-NeRF~\cite{barron2021mip} and Mip-NeRF360~\cite{barron2022mip} are designed specifically for unbounded outdoor scenes, our method performs favorably in all metrics. 

Urban-NeRF~\cite{rematas2022urban} uses depth cues from LiDAR in a NeRF model to reconstruct urban scenes.
In contrast, we leverage LiDAR as a geometric prior in the proposed Gaussian models to achieve more effective large-scale scene reconstruction.
Our method performs favorably against S-NeRF~\cite{xie2023s} and SUDS~\cite{turki2023suds}, where both decompose the scene into static background and dynamic objects and construct the scene with the help of LiDAR.
Compared to EmerNeRF~\cite{yang2023emernerf}, which applies a spatial-temporal representation for dynamic driving scenes using flow fields, our method achieves state-of-the-art results in all metrics, eliminating the need to estimate scene flow.
For Gaussian-based approaches, our method enhances the performance of our baseline method, 3DGS~\cite{kerbl20233d}, on large-scale scenes in all metrics. 

We show qualitative evaluation results on challenging nuScenes driving scenes.
For multicamera surround view synthesis, as shown in Figure~\ref{temporal_compare}, our method enables the generation of photorealistic rendering images and ensures view consistency across multicameras. 
Meanwhile, EmerNeRF~\cite{yang2023emernerf} and 3DGS~\cite{kerbl20233d} do not perform well in challenging regions, exhibiting undesirable visual artifacts such as ghosting, dynamic object disappearance, loss of plant texture details, lane markings, and blurring in distant scenes.  

We next demonstrate the reconstruction results for dynamic temporal scenes.
Our method accurately models dynamic objects within large-scale scenes, mitigating issues such as loss, ghosting, or blurring of these dynamic elements. 
The proposed model consistently constructs dynamic objects over time, despite their relatively high speed of movement.
As depicted in Figure~\ref{temporal_compare}, other approaches~\cite{yang2023emernerf, kerbl20233d} are inadequate for rapidly moving dynamic objects.

\begin{figure*}[tp]
  \centering
  \includegraphics[width=0.95\linewidth]{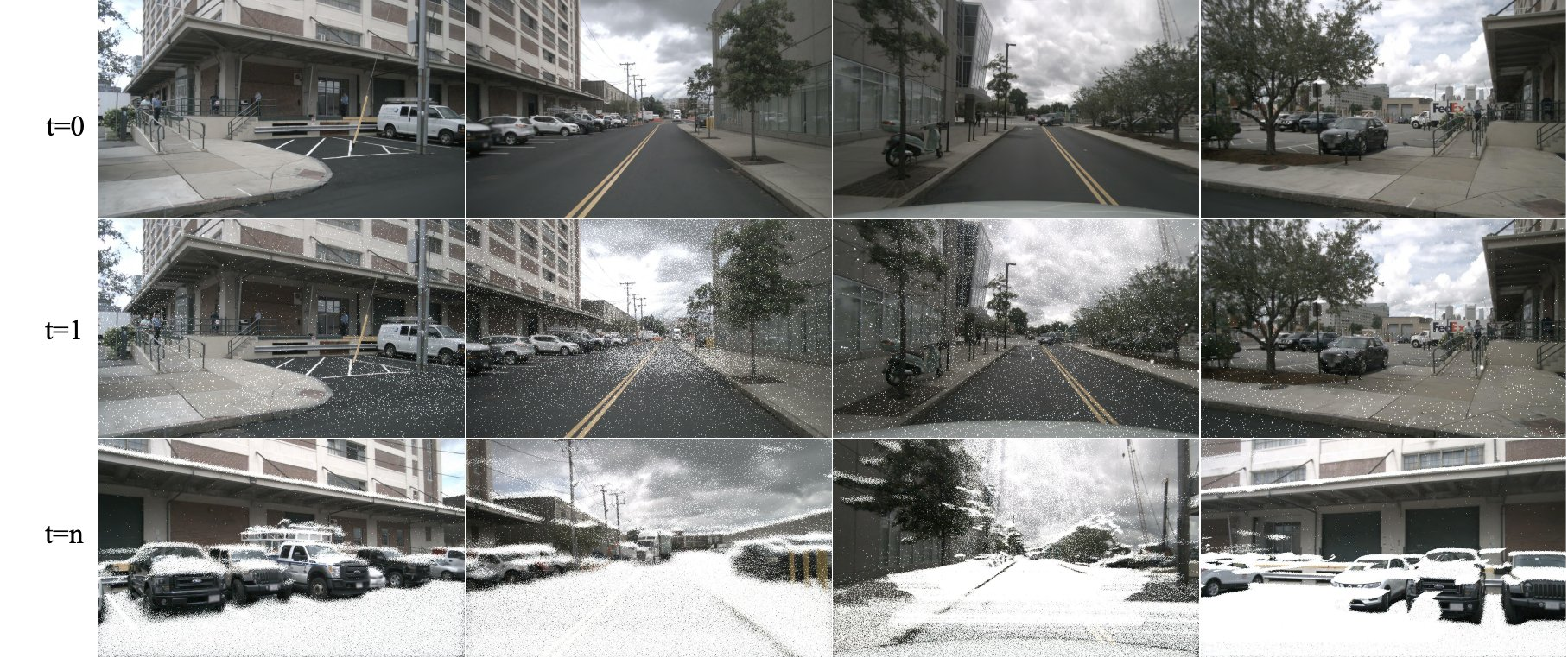}
  \caption{\textbf{Dynamic Simulation of Snow.} We first add dynamic particles at each time step. Secondly, we estimate the surface normal to obtain the particle deposition position and specify the particle motion trajectory. Our method can generate snowy scenes that follow physical principles.}
  \label{fig:snow_dynamic}
  \vspace{-6pt}
\end{figure*}

\begin{figure*}[tp]
  \centering
  \includegraphics[width=0.95\linewidth]{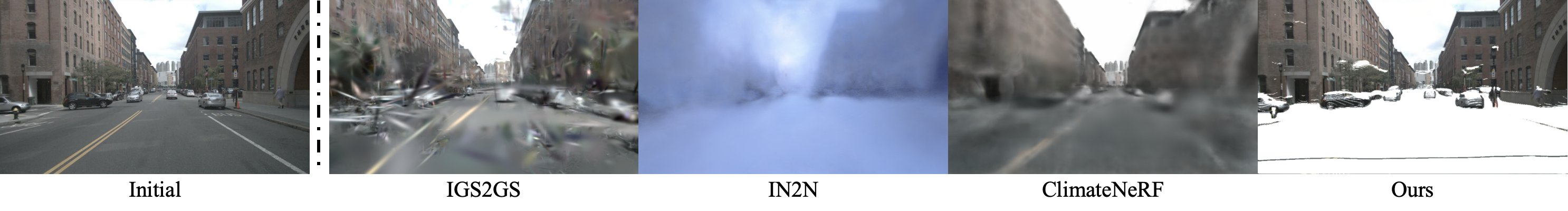}
  \caption{\textbf{Qualitative Comparison of Weather Simulation.} We compare our method with IN2N, IGS2GS, and ClimateNeRF for weather simulation on 4D driving scenes from the nuScenes dataset. \ourmethod{} delivers realistic and coherent weather editing while maintaining high efficiency.}
  \label{fig:3D_cmp}
\end{figure*}

\subsubsection{Comparisons of mono-view on KITTI-360}
To further validate the effectiveness of our method in the setting of a monocular driving scene, we conduct experiments with the KITTI-360 data set with comparisons to existing state-of-the-art approaches, including NeRF~\cite{mildenhall2021nerf}, Mip-NeRF360~\cite{barron2022mip}, Point-NeRF~\cite{xu2022point}, NSG~\cite{ost2021neural}, SUDS~\cite{turki2023suds}, and DNMP~\cite{lu2023urban}. As shown in Table~\ref{KITTI-360}, our method performs favorably in monocular driving scenes against other models.

\subsection{Editing Results and Comparisons}
We first demonstrate our editing results on the nuScenes dataset for multiple tasks. Compared with state-of-the-art 2D and 3D editing methods, our approach achieves superior visual realism and better quantitative consistency.

To support flexible editing on driving scenes, we additionally created a 3D Gaussian foreground bank containing specialized driving scene objects. This foreground bank is critically important for autonomous driving simulation and model validation.

\subsubsection{Qualitative Results and Comparisons}
We perform training-free editing of the reconstructed nuScenes data by \ourmethod{} in three domains: texture, weather, and object manipulation. The comprehensive results are shown in Fig~\ref{fig:res}, showcasing the ability of \ourmethod{} to perform various editing operations in dynamic driving scenarios.

\begin{figure*}[tp]
  \centering
  \includegraphics[width=.95\linewidth]{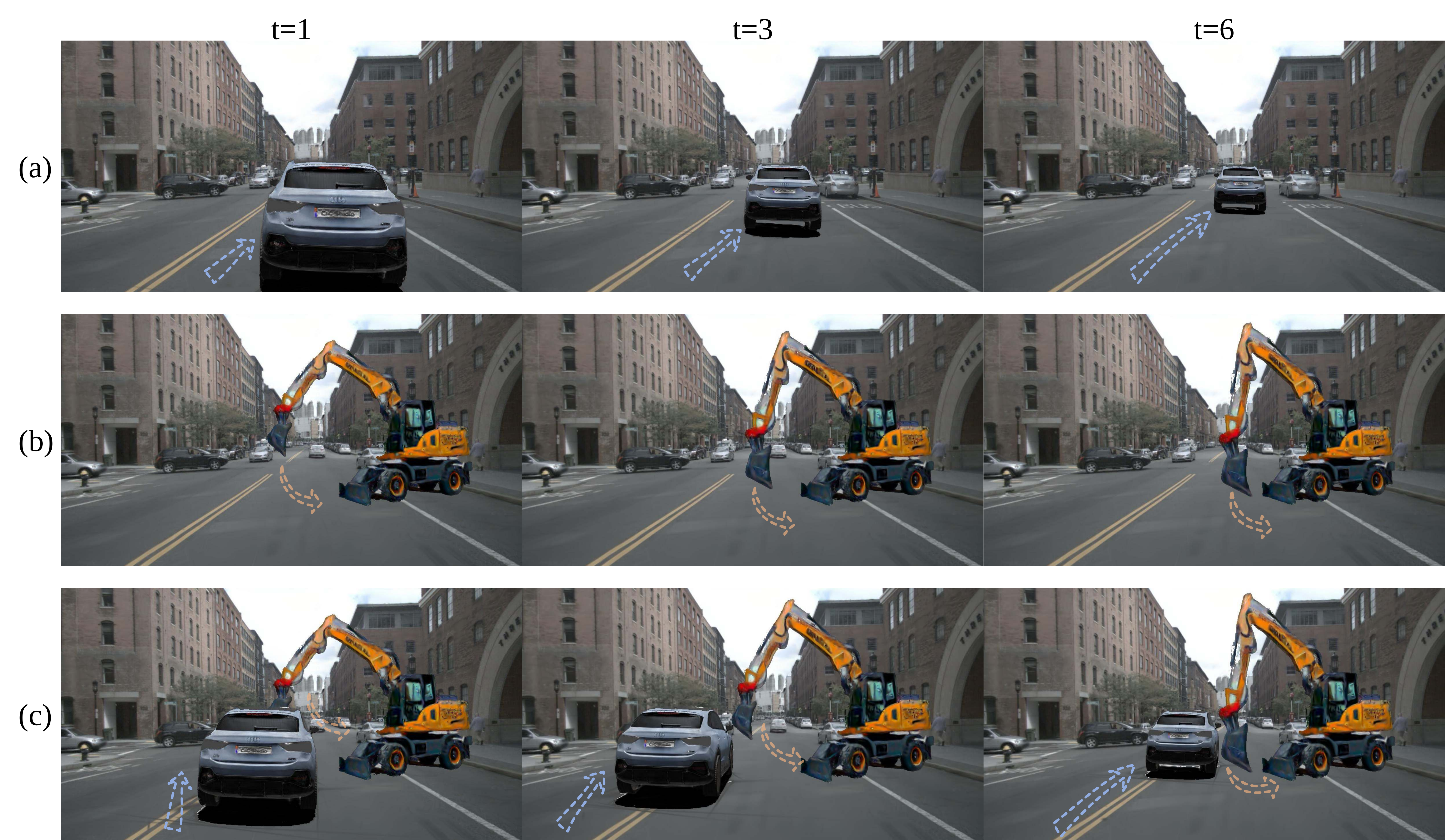}
  \caption{\textbf{Dynamic Object Insertion and Scene Integration.} We generate deformable foreground objects and seamlessly insert them into the scene. \textbf{(a)} A rigid car is added. \textbf{(b)} A generated 4D excavator is integrated. \textbf{(c)} Multiple objects are inserted with correct occlusion. The results demonstrate the effectiveness of our method in achieving natural object insertion with realistic spatial interactions. 
  }
  \label{fig:dynamic_res}
  \vspace{-4pt}
\end{figure*}

For weather editing, we achieve realistic effects through particle-based simulation, as described in Section~\ref{3.1:creation}. For snow simulation particularly, we add snow particle Gaussians at each timestep and estimate surface normals to determine deposition locations. This produces realistic snow accumulation, as shown in Figure~\ref{fig:snow_dynamic}. 
For object manipulation, we first obtain actors using 3D reconstruction or 4D generation methods such as DreamGaussian4D~\cite{ren2023dreamgaussian4d}. This allows for the insertion of dynamic objects, including non-rigid bodies such as humans and animals. By adapting the deformation module to foreground contexts, we achieve flexible and diverse dynamic object integration. Additionally, we employ LLM-based trajectory prediction to obtain the trajectory of inserted objects. The results of dynamic objects insertion are shown in Figure~\ref{fig:dynamic_res}.

Fig~\ref{fig:3D_cmp} provides a performance comparison with existing 3D editing approaches. While InstructNeRF2Nerf~\cite{haque2023instruct} and InstructGS2GS~\cite{vachhainstruct} employ diffusion models for iterative 3D scene editing across multiple tasks, they exhibit limitations in maintaining photorealism and view consistency. ClimateNeRF~\cite{li2023climatenerf} specializes in particle-level weather editing through surface normal computations, but its application lacks generalizability to other editing tasks and remains constrained to static environments. Our method addresses these limitations while achieving high-quality results across all editing tasks.

\begin{figure*}[tbp]
  \centering
  \includegraphics[width=0.95\linewidth]{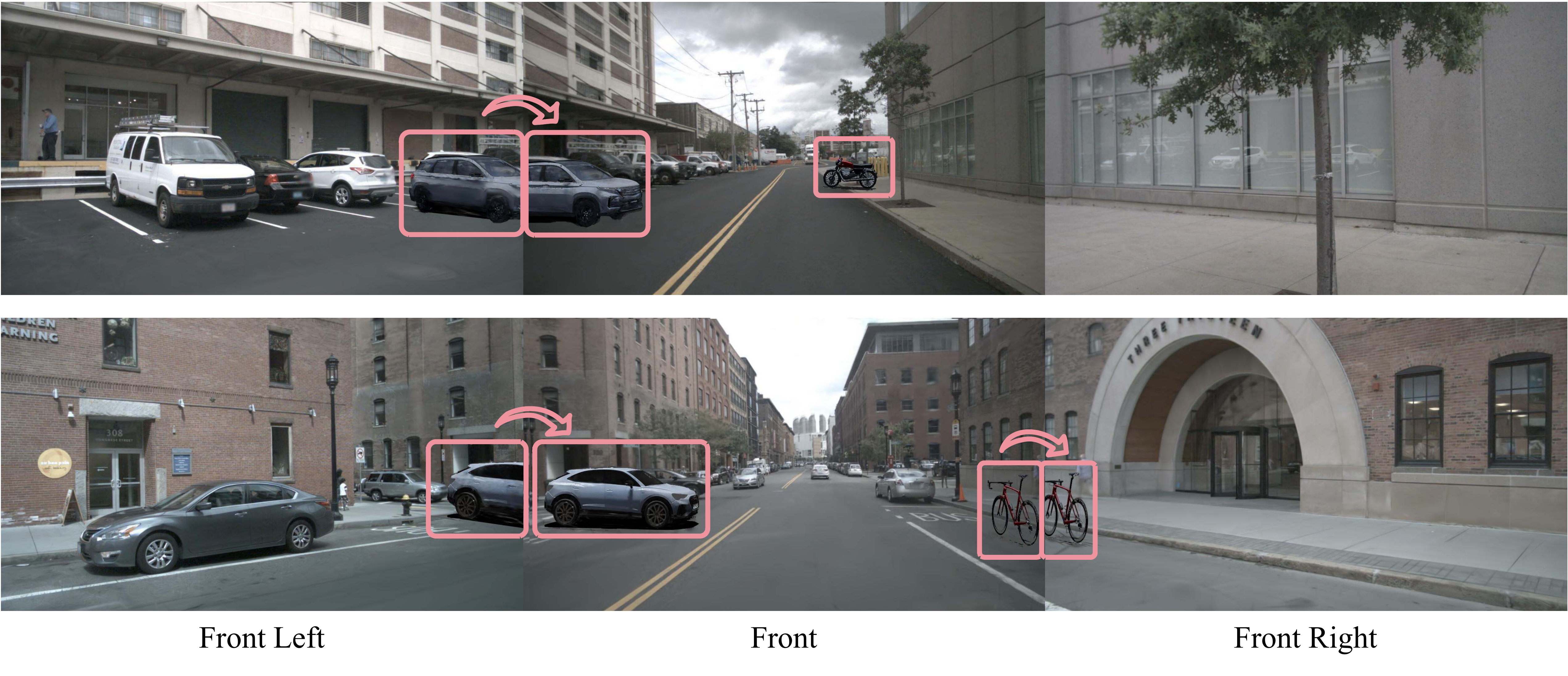}
  \caption{\textbf{Multiple Objects Insertion with Foreground Bank.} We insert distinct objects from the foreground bank into each scene across various viewpoints. The results exhibit strong cross-view consistency in both geometry and appearance, highlighting the robustness of our method for 3D object insertion in complex driving scenes.}
  \label{fig:add_scene}
  \vspace{-6pt}
\end{figure*}

\begin{table*}[ht]
\centering
\caption{\textbf{Quantitative Comparison of \ourmethod{} with State-of-the-Art Image-Editing Approaches.} We compare our method with existing state-of-the-art techniques on the nuScenes dataset across various editing tasks. The results show competitive performance, with improved quantitative metrics and enhanced editing quality.
}
\begin{minipage}[h]{0.4\textwidth}
\centering
        \begin{tabular}{cccc}
            \toprule
        \textbf{Methods} & \textbf{Type} & $\textbf{CLIP}_{dir} \uparrow$ \\    
        \hline
        Paint-by-Example~\cite{yang2023paint}    & 2D & \cellcolor{tabsecond}0.0282    \\
        AnyDoor~\cite{chen2024anydoor} & 2D  & 0.0027  \\
        \hline
        \textbf{Ours}      & 3D                  & \cellcolor{tabfirst}0.0866     \\
        \hline
        \end{tabular}
    \caption*{(a) Object Insertion Task.}
\end{minipage}
\begin{minipage}[h]{0.4\textwidth}
\centering
    \begin{tabular}{ccc}
        \toprule
    \textbf{Methods} & \textbf{Type} & \textbf{LPIPS} $\downarrow$  \\    
    \hline
    SD-inpainting~\cite{rombach2022high}   & 2D    & 0.3435 \\
    InstructDiffusion~\cite{geng2024instructdiffusion} & 2D & \cellcolor{tabsecond}0.3271  \\
    \hline
   \textbf{Ours}             & 3D           & \cellcolor{tabfirst}0.3286   \\
    \hline
    \end{tabular}
    \caption*{(b) Object Removal Task.}
\end{minipage}

\begin{minipage}[h]{0.4\textwidth}
\centering
    \begin{tabular}{cccc}
            \toprule
    \textbf{Methods} & \textbf{Type} & $\textbf{CLIP}_{dir} \uparrow$ \\    
    \hline
    InstructDiffusion~\cite{geng2024instructdiffusion} & 2D & 0.0040 \\ 
    InstructPix2Pix~\cite{brooks2023instructpix2pix}  & 2D & 0.2241 \\   
    FreePromptEditing~\cite{liu2024towards} & 2D & 0.1709 \\
    UltraEdit~\cite{zhao2024ultraedit} & 2D & \cellcolor{tabsecond}0.2292 \\
    \hline
    \textbf{Ours}       & 3D                 & \cellcolor{tabfirst}0.2462     \\
    \hline
    \end{tabular}
    \caption*{(c) Weather Edit Task.}
\end{minipage}
\begin{minipage}[h]{0.4\textwidth}
\centering
    \begin{tabular}{cccc}
            \toprule
    \textbf{Methods} & \textbf{Type} & $\textbf{CLIP}_{dir} \uparrow$ \\    
    \hline
    Paint-by-Example~\cite{yang2023paint} & 2D  & 0.0940  \\
    AnyDoor~\cite{chen2024anydoor} & 2D & \cellcolor{tabsecond}0.1358  \\
    SD-inpainting~\cite{rombach2022high} & 2D  & 0.0493 \\
    UltraEdit~\cite{zhao2024ultraedit} & 2D & 0.0427 \\
    \hline
    \textbf{Ours}     & 3D                   & \cellcolor{tabfirst}0.2019     \\
    \hline
    \end{tabular}
    \caption*{(d) Texture Edit Task.}
\end{minipage}
\label{compare_SOTA_2}
\vspace{-6mm}
\end{table*}

\subsubsection{Quantitative Results and Comparisons}

To evaluate the consistency and realism of our editing approach, we compare \ourmethod{} with state-of-the-art 3D and 2D editing techniques. 

For 3D scene editing, we compare with ClimateNeRF~\cite{li2023climatenerf}, IN2N~\cite{haque2023instruct}, and IGS2GS~\cite{vachhainstruct} in terms of task diversity, processing time, and CLIP-direction similarity~\cite{brooks2023instructpix2pix}. As shown in Table~\ref{table:efficiency}, \ourmethod{} performs favorably against all other methods in terms of diversity, efficiency, and text-aligned consistency. In particular, the editing time of \ourmethod{} is typically within 3$\sim$10 minutes for scenes from the NuScenes dataset, significantly lower than that of other 3D editing models that require long training time.

To evaluate the performance of \ourmethod{} on single-view editing, we also compare it with 2D editing methods~\cite{chen2024anydoor,yang2023paint,rombach2022high,liu2024towards,brooks2023instructpix2pix,geng2024instructdiffusion,suvorov2022resolution} across different tasks, as shown in Table~\ref{compare_SOTA_2}.
For texture modification and object insertion, we compare with inpainting methods~\cite{chen2024anydoor,yang2023paint,rombach2022high}. While AnyDoor~\cite{chen2024anydoor} and Paint-by-Example~\cite{yang2023paint} utilize 2D images for conditional editing, they produce inconsistent perspective relationships and poor consistency with the condition image. SD-Inpainting~\cite{rombach2022high} takes text prompts and 2D masks as input, but suffers from limited performance and controllability.
For weather simulation, we evaluate text-guided editing methods~\cite{liu2024towards,brooks2023instructpix2pix,geng2024instructdiffusion}. Although FreePromptEditing~\cite{liu2024towards}, InstructPix2Pix~\cite{brooks2023instructpix2pix}, and InstructDiffusion~\cite{geng2024instructdiffusion} exhibit good text understanding, their results often lack physical plausibility—for instance, snow is rendered merely as a stylistic change rather than as accumulated precipitation.
InstructDiffusion~\cite{geng2024instructdiffusion} editing results in these weather scenes are less realistic. 
For object removal, we evaluate inpainting and text-guided methods~\cite{rombach2022high,geng2024instructdiffusion,suvorov2022resolution}. SD-Inpainting~\cite{rombach2022high} and InstructDiffusion~\cite{geng2024instructdiffusion} leave residual artifacts, while LaMa~\cite{suvorov2022resolution} introduces visible inconsistencies in scene restoration.

We evaluate editing consistency using CLIP direction similarity metric for texture, weather editing and object insertion. For object removal, we evaluate the quality using LPIPS and FID (as shown in SPIn-NeRF~\cite{mirzaei2023spin}). \ourmethod{} achieves superior performance on all tasks.

\subsubsection{3D Gaussian Foreground Bank for Driving Scenes}
We construct a comprehensive 3D Gaussian foreground bank containing various traffic elements: vehicles, bicycles, motorcycles, pedestrians, animals, and static objects such as signs and traffic cones. Figure~\ref{fig:add_scene} shows our foreground bank and insertion results.

\noindent \textbf{Online Model Reconstruction.} We collect 3D models (pedestrians, vehicles, etc.) from online sources and Chatsim~\cite{wei2024editable}, then reconstruct them using 3DGS~\cite{kerbl20233d}. For each model, we render 360\textdegree views in Blender and perform 3DGS reconstruction with COLMAP~\cite{schonberger2016structure}. We adjust lighting using environment maps extracted from nuScenes.

\noindent \textbf{Sparse Reconstruction of vehicles from nuScenes.} We efficiently sparse reconstruct vehicles from nuScenes using SplatterImage~\cite{szymanowicz2024splatter}. Each vehicle requires about 2$\sim$4 reference images for Gaussian reconstruction.

\noindent \textbf{Image-based Object Generation.} To expand our dataset, we generate 3D objects with image inputs. We first extract clean object images with SAM~\cite{zou2023segment}. Subsequently, we create static and dynamic 3D models using dreamgaussian~\cite{tang2023dreamgaussian} and dreamgaussian4d~\cite{ren2023dreamgaussian4d} for few-shot 3D generation, enabling the creation of static and dynamic objects with high fidelity and efficiency.
\begin{figure*}[t]
  \centering
  \includegraphics[width=0.9\linewidth]{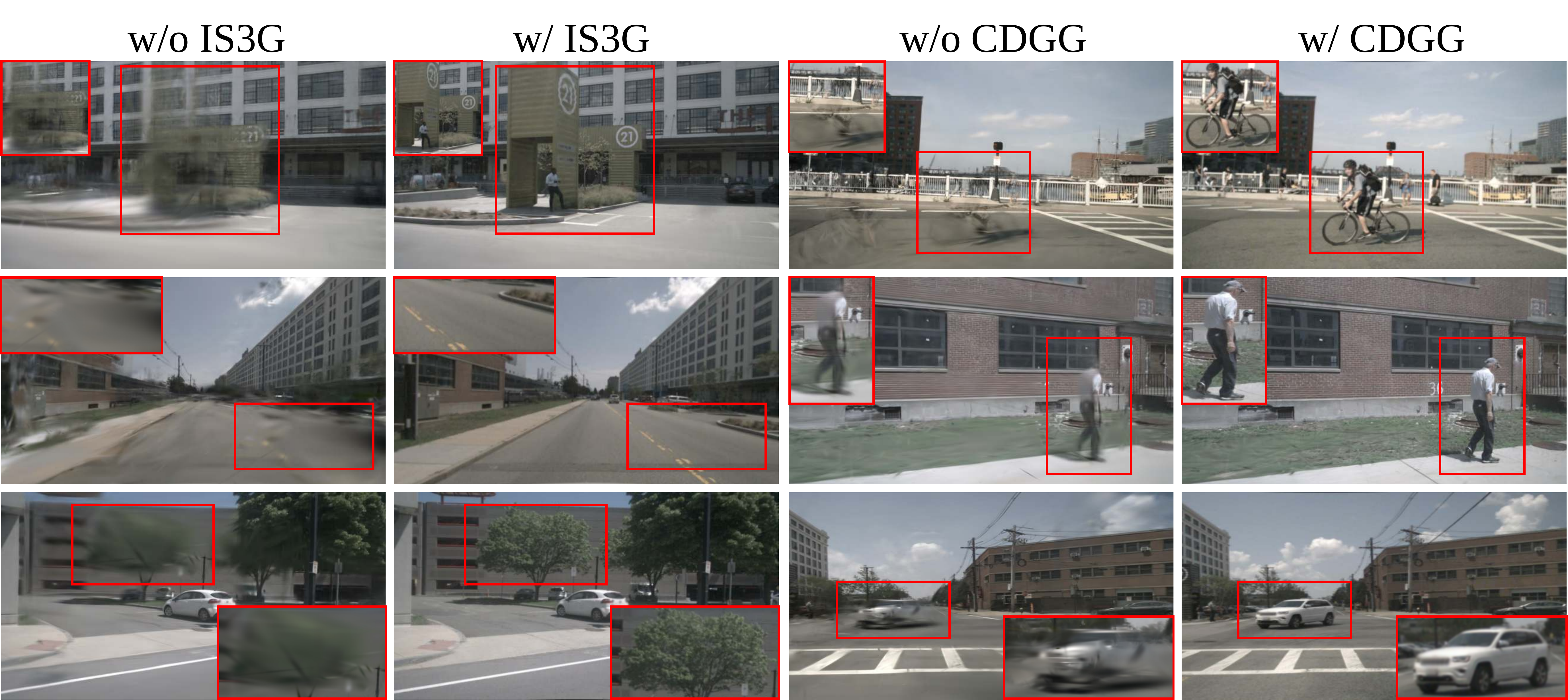}
  \caption{\textbf{Rendering with or w/o the Incremental Static 3D Gaussians (IS3G) and Composite Dynamic Gaussian Graph (CDGG).} IS3G ensures good geometry and topological integrity for static backgrounds in large-scale driving scenes. CDGG enables the reconstruction of dynamic objects at arbitrary speeds in driving scenes (e.g., vehicles, bicycles, and pedestrians).}
  \label{staticdynamic-effect}
 \end{figure*}

\begin{figure}[tp]
  \centering
  \includegraphics[width=0.95\linewidth]{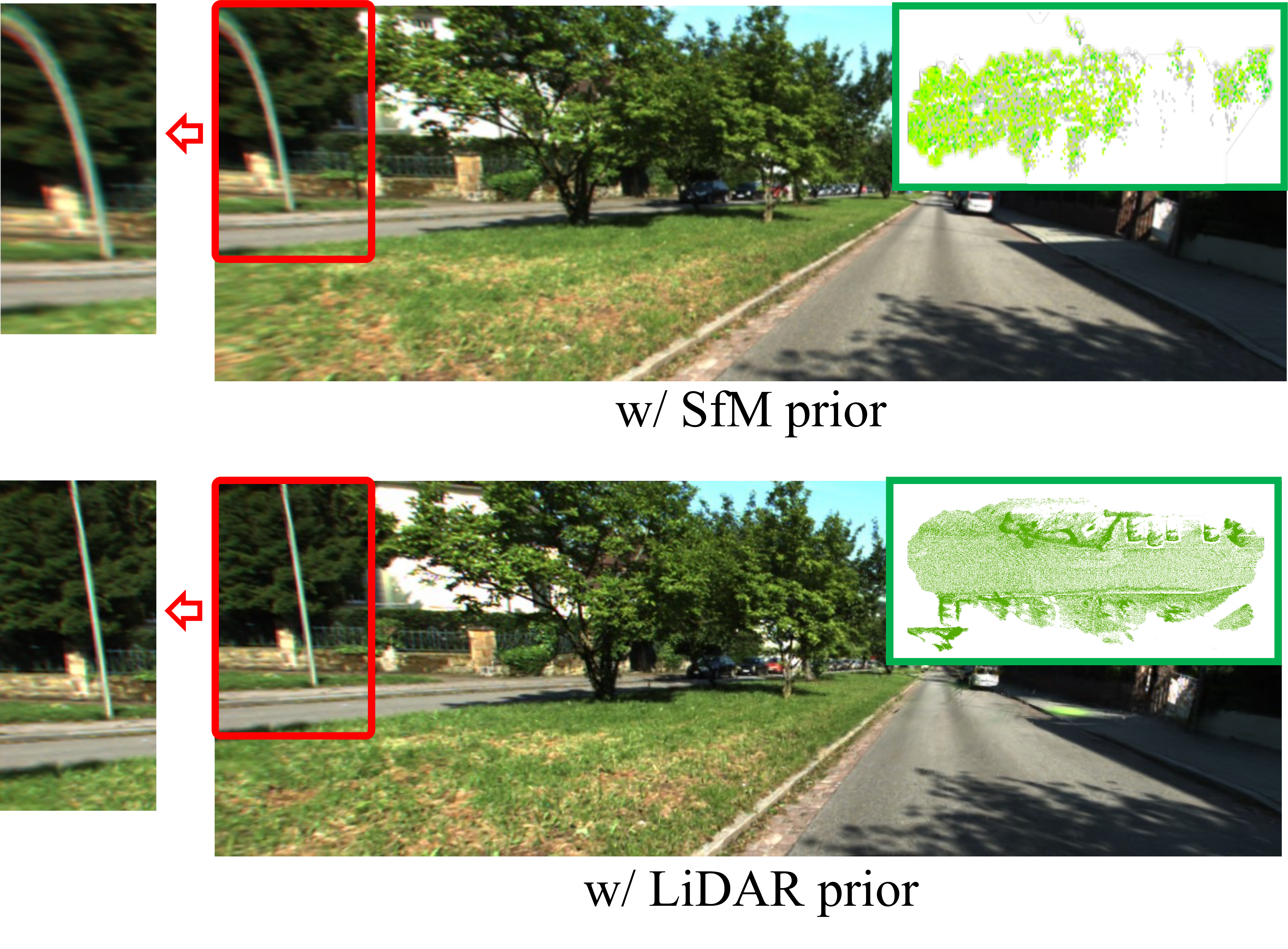}
  \caption{\textbf{Visualization comparison using different initialization methods on KITTI-360.} Compared to initialization with SfM points~\cite{kerbl20233d}, using LiDAR prior allows Gaussians to restore more accurate geometric structures in the scene.}
  \label{fig:effect-LiDAR}
 \end{figure}

\subsection{Ablation Study}

\subsubsection{Initialization prior for Gaussians} Comparative experiments are conducted to analyze the effect of different priors and initialization methods on the Gaussian model. The original 3DGS provides two initialization modes: randomly generated points and SfM points computed by COLMAP~\cite{schonberger2016structure}. We additionally offer two other approaches: point clouds from a pre-trained NeRF model and points generated with LiDAR prior.

Meanwhile, to analyze the effect of point cloud quantity, we down-sample the LiDAR to 600K and apply adaptive filtering (1M) to control the number of generated LiDAR points. We also set different maximum thresholds for randomly generated points (600K and 1M). Here, SfM-600K$\pm$ 20K represents the number of points computed by COLMAP, NeRF-1M$\pm$20K denotes the total points generated by the pre-trained NeRF model, and LiDAR-2M$\pm$20k refers to the original quantity of LiDAR points.

As shown in Table~\ref{LiDAR_prior}, randomly generated points lead to the worst results as they lack any geometric prior. Initializing with SfM points also cannot adequately recover the scene's precise geometries due to the sparse points and intolerable structural errors. Leveraging point clouds generated from a pre-trained NeRF model provides a relatively accurate geometric prior, but there are still noticeable outliers. For the model initialized with the LiDAR prior, although downsampling results in loss of geometric information in some local regions, it still retains relatively accurate structural priors, thus surpassing SfM (Figure~\ref{fig:effect-LiDAR}). 
We note that the experimental results do not change linearly with increasing LiDAR point quantities. 
This can be attributed to that overly dense points store redundant features that interfere with the optimization of the Gaussian model.

\begin{table}[!t]
  \centering
  \caption{
  {\textbf{Effect of different initialization methods on the Gaussian model.} LiDAR-600K $\dagger$ denotes downsampling the original LiDAR data to a corresponding point cloud magnitude. LiDAR-1M  $\ddagger$ denotes denoising and removing outliers in LiDAR points, which is used in our method.}}
    \footnotesize
  \centering
  \setlength{\tabcolsep}{1.4mm}{
  {
    \begin{tabular}{cccc}
   \toprule
    \textbf{Methods} & \textbf{PSNR $\uparrow$} & \textbf{SSIM $\uparrow$} & \textbf{LPIPS $\downarrow$} \\
    \hline 
    Random-600K   & 22.18     & 0.653     & 0.424  \\
    Random-1M     & 22.23     & 0.653     & 0.421  \\
    SfM-600K      & 28.36     & 0.851     & 0.256  \\
    NeRF-1M       & 28.51     & 0.858     & 0.251  \\
    \hline
    LiDAR-600K $\dagger$      & 28.49     & 0.854     & \cellcolor{tabsecond} 0.245 \\
    LiDAR-1M  $\ddagger$      & \cellcolor{tabsecond} 28.74     & \cellcolor{tabsecond} 0.865     & \cellcolor{tabfirst} 0.237  \\
    LiDAR-2M                  & \cellcolor{tabfirst} 28.78     & \cellcolor{tabfirst} 0.867     & \cellcolor{tabfirst} 0.237  \\
    \hline
    \end{tabular}%
    }
    }
  \label{LiDAR_prior}%
  \aftertab
\end{table}%

\subsubsection{Effectiveness of Model Component}
We analyze the contribution of each module of the proposed model. As shown in Table~\ref{Ablation_Study} and Figure~\ref{staticdynamic-effect}, the Composite Dynamic Gaussian Graph module plays a crucial role in reconstructing dynamic driving scenes, while the Incremental Static 3D Gaussians module enables high-quality large-scale background reconstruction. These two novel modules significantly enhance the modeling quality of complex driving scenes. Regarding the proposed loss functions, the ablation results indicate that both $L_{TSSIM}$ and $L_{Robust}$ significantly improve the rendering quality, improve texture details and remove artifacts. In addition, $L_{LiDAR}$ from LiDAR prior helps Gaussians achieve better geometric priors. The experimental results also demonstrate that \ourmethod{} performs well even without the prior LiDAR, demonstrating strong robustness for various initialization methods.

\begin{table}[t]
  \centering
  \caption{
  {\textbf{Effect of each module in our proposed method.} IS3G is short for the Incremental Static 3D Gaussians module, and CDGG is short for the Composite Dynamic Gaussian Graph module.}}
    \footnotesize
  \centering
  \setlength{\tabcolsep}{1.4mm}{
  {
    \begin{tabular}{cccc}
 \toprule
    \textbf{Model} & \textbf{PSNR $\uparrow$} & \textbf{SSIM $\uparrow$} & \textbf{LPIPS $\downarrow$} \\
    \hline 
    w/o IS3G             & 27.72     & 0.771     & 0.295  \\
    w/o CDGG             & 26.97     & 0.752     & 0.306  \\
    w/o $L_{TSSIM}$      & 27.88     & 0.783     & 0.280  \\
    w/o $L_{Robust}$     &  28.05     &  0.814     & 0.271  \\
    w/o $L_{LiDAR}$      & \cellcolor{tabsecond} 28.45     & \cellcolor{tabsecond} 0.854     & \cellcolor{tabsecond} 0.248  \\
    Ours-S               &  28.36 &  0.851 &  0.256  \\
    Ours-L               & \cellcolor{tabfirst} 28.74  & \cellcolor{tabfirst} 0.865 & \cellcolor{tabfirst} 0.237  \\
    \hline
    \end{tabular}%
    }
    }
  \label{Ablation_Study}%
  \aftertab
\end{table}%

\section{Conclusion}
We introduce \ourmethod{}, a framework for reconstructing and editing large-scale dynamic autonomous driving scenes. Our approach progressively models the static background using incremental static 3D Gaussians and captures multiple moving objects through a composite dynamic Gaussian graph. By leveraging LiDAR priors, we achieve accurate geometric structures and robust multi-view consistency, significantly enhancing the quality of scene reconstruction. \ourmethod{} facilitates training-free editing for tasks such as texture modification, weather simulation, and object manipulation, enabling the generation of realistic and diverse driving scenes. Experimental results on datasets such as nuScenes and KITTI-360 demonstrate that our framework achieves state-of-the-art performance in both reconstruction and editing tasks, enabling high-quality surrounding view synthesis and dynamic scene editing. %

\bibliography{main}
\bibliographystyle{IEEEtran}





\end{document}